\documentclass[journal]{IEEEtran}

\usepackage{graphicx}
\usepackage{tabularx}
\usepackage{svg}
\usepackage{amsmath,amssymb}
\usepackage{booktabs}
\usepackage{array}
\usepackage{multirow}
\usepackage[table]{xcolor}
\usepackage{url}
\usepackage{listings}
\usepackage[hidelinks]{hyperref}

\usepackage{orcidlink}

\newcommand{\todo}[1]{}

\begin{document}
\bstctlcite{IEEEexample:BSTcontrol}

\title{LATCH: LLM-Assisted Timing Closure Harness for Post-Route FPGA Optimisation}

\author{
  Shu~Gu \textsuperscript{\orcidlink{0009-0008-7617-3350}},
  Matthew~Youngman \textsuperscript{\orcidlink{0009-0007-0981-3082}}, \IEEEmembership{Graduate~Student~Member,~IEEE},
  Yuqin~Zhao \textsuperscript{\orcidlink{0000-0001-7943-1433}}, \\
  Cristian~Sestito \textsuperscript{\orcidlink{0000-0002-7731-0002}}, \IEEEmembership{Member,~IEEE},
  Ali~Siddique \textsuperscript{\orcidlink{0000-0002-7688-6878}},
  Ramesh~Krishnamurthy \textsuperscript{\orcidlink{0000-0002-1098-3587}},
  Danial~Chitnis \textsuperscript{\orcidlink{0000-0003-0916-9921}},
  and~Themis~Prodromakis \textsuperscript{\orcidlink{0000-0002-6267-6909}}, \IEEEmembership{Senior Member,~IEEE}

  \thanks{This work was supported by the Engineering and Physical Sciences Research Council (EPSRC) AI Hub for Productive Research and Innovation in eLectronics (APRIL) under Grant No. EP/Y029763/1, and by the Royal Academy of Engineering (RAEng) Chair in Emerging Technologies under Grant No. CiET1819/2/93. \textit{(S. Gu and M. Youngman contributed equally to this work. Corresponding author: Matthew Youngman)}.

  S. Gu, M. Youngman, Y. Zhao, C. Sestito, A. Siddique, R. Krishnamurthy and T. Prodromakis are with the Centre for Electronics Frontiers, Institute for Integrated Micro and Nano Systems, School of Engineering, The University of Edinburgh, U.K. (email: m.j.youngman@sms.ed.ac.uk).

  D. Chitnis is with the Institute for Integrated Micro and Nano Systems, School of Engineering, The University of Edinburgh, U.K.}
}

\maketitle

\begin{abstract}
Post-route FPGA timing closure remains expert-dependent and computationally expensive. We present LATCH (LLM-Assisted Timing Closure Harness), a two-layer framework that separates large language model (LLM) orchestration from deterministic FPGA implementation and evaluation. Across sixteen designs and seven LLMs, a single-iteration study compares three agent control granularities -- expert-written recipes, direct EDA operations and LLM-constructed procedures -- to ask how much control an agent needs per attempt. Functional-pass rates are 98.2\%, 71.4\% and 17.9\%, respectively; yet direct operations give slightly larger 3.4\% median $F_{\max}$ gains when passing, versus 3.2\% for recipes. This exposes a trade-off between execution reliability and timing improvement. Informed by this trade-off, a multi-iteration study fixes the action tier at expert recipes to ask how repeated attempts are best organised, comparing a fitted deterministic heuristic with sequential LLM reasoning, branching exploration and knowledge base-guided exploration within 12 checkpoints per run. Branching raises mean maximum-frequency improvement from 10.5\% (sequential) to 16.0\%, and the knowledge base to 20.3\%, improving 15 of 16 designs. Characterisation across 2,421 recipe attempts profiled their gain and validation rates, showing hard-block re-placement gives the largest cumulative mean gain at 41.4~MHz.
\end{abstract}


\section{Introduction}
\label{sec:introduction}

\newcolumntype{Y}{>{\raggedright\arraybackslash}X}

Timing closure seeks to meet a design's target frequency by diagnosing failing paths and modifying placement and routing (P\&R), physical optimisation, or constraints to improve $F_{\max}$. This remains a critical but expert-dependent stage because the appropriate intervention is determined by design-specific physical bottlenecks \cite{leaps2024} and their interactions across many available optimisation options~\cite{datuner2017}. Each trial can require a full P\&R run, with large circuits taking hours or even days~\cite{datuner2017, flowtune2023}. An autonomous timing-optimisation method could reduce this cost by selecting interventions from physical evidence.

Machine learning (ML) has been applied across the EDA flow to improve engineering productivity, either by predicting physical outcomes earlier in the flow~\cite{imlg2024} or by selecting tool settings more intelligently~\cite{huang2021mleda}. Within FPGA timing optimisation, predictive models estimate post-route net delay during placement~\cite{martin2021}, calibrate delay inside differentiable timing-driven placement~\cite{lin2025}, and weight global placement by predicted criticality~\cite{jiang2026}. By acting inside the placement algorithm, these methods improve the next implementation but cannot choose a local repair for an existing routed checkpoint. Autotuners instead learn which tool settings or optimisation sequences to try: InTime~\cite{intime2015} and LAMDA~\cite{lamda2019} rank tool-parameter combinations, DATuner~\cite{datuner2017} allocates trials across parameter-space partitions, and FlowTune builds optimisation sequences stage by stage, favouring sub-sequences with better measured results~\cite{flowtune2023}. These autotuners remove manual tuning, but every candidate costs at least a partial implementation rerun and each model is learned within a single design's campaign. As a result, these methods remain bound to costly iterations and generalise poorly to unseen designs. Where a learned placement policy has reportedly generalised, it relied on pre-training over earlier designs~\cite{mirhoseini2021}, with cross-design generalisation as an open research problem~\cite{huang2021mleda}.

Large language models (LLMs) target these learning and generalisation limitations because they can reason over textual and structured evidence~\cite{Pan_2025}, compare it with tool documentation or prior examples~\cite{timingagent,rsr}, and select actions from tool feedback without fitting a new selector for each design~\cite{eesizer2026}. This is attractive for timing optimisation, where useful evidence appears in timing reports, logs, constraint files and optimisation histories~\cite{timingagent}. Physical implementation nevertheless exposes a hardware-specific failure mode termed the \textit{syntax trap}, where LLM outputs can appear plausible as text, scripts or RTL code while remaining physically ineffective, unsafe or disconnected from downstream implementation behaviour~\cite{llms_in_digital_eda}. This trap is severe in backend design because public physical-design data is scarce, locked behind licenses, and expensive to generate ~\cite{edacorpus2024,fu2025hwdebug}. Physical-design benchmarks confirm the gap, as agents degrade on geometric reasoning and power, performance and area (PPA) trade-offs~\cite{posteda2026}, and on tool-centric, long-horizon workflows~\cite{pdagentbench2026}.

\begin{table*}[t]
\centering
\caption{Selected learning-based approaches to backend timing and physical optimisation. Groups, from top: ML-based, LLM-based, and this work (shaded). Entries refer to the reported experiments.}
\label{tab:prior_work}
\begingroup
\fontsize{7.8}{9.3}\selectfont
\setlength{\tabcolsep}{2.5pt}
\renewcommand{\arraystretch}{1.02}
\begin{tabularx}{\textwidth}{@{}>{\raggedright\arraybackslash}p{29mm}>{\raggedright\arraybackslash}X>{\centering\arraybackslash}p{16mm}>{\centering\arraybackslash}p{17mm}>{\centering\arraybackslash}p{17mm}>{\centering\arraybackslash}p{17mm}>{\raggedright\arraybackslash}p{26mm}>{\centering\arraybackslash}p{17mm}@{}}
\toprule
\textbf{Work} & \textbf{Target / Input} & \textbf{Designs} & \textbf{Timing-feedback loop} & \textbf{Physical modification} & \textbf{Knowledge reuse} & \textbf{Search} & \textbf{Control-level study} \\
\midrule
InTime (2015)~\cite{intime2015} & FPGA / RTL & 6 & $\checkmark$ & $\checkmark^{\mathrm{F}}$ & $\checkmark$ & Parallel batches & NR \\
DATuner (2017)~\cite{datuner2017} & FPGA / RTL & 15 (10+5) & $\checkmark$ & $\checkmark^{\mathrm{F}}$ & $\checkmark$ & Adaptive batches & NR \\
LAMDA (2019)~\cite{lamda2019} & FPGA / RTL & 5 & $\checkmark^{\mathrm{E}}$ & $\checkmark^{\mathrm{F,E}}$ & $\checkmark$ & Multi-stage batches & NR \\
\midrule
Nexus (2025)~\cite{nexus} & FPGA / RTL + constraints & 14 & $\checkmark$ & $\checkmark^{\mathrm{F}}$ & NR & Sequential refinement & NR \\
ORFS-agent (2025)~\cite{orfs} & ASIC / RTL + PDK & 3 (2 nodes) & $\checkmark$ & $\checkmark^{\mathrm{F}}$ & $\checkmark$ & Parallel batches & NR \\
FluxEDA (2026)~\cite{fluxeda} & ASIC / routed design & 1 (ECO case) & $\checkmark$ & $\checkmark^{\mathrm{D}}$ & $\checkmark$ & Sequential + rollback & NR \\
PACT (2026)~\cite{pact} & FPGA / routed DCP & 35 DCPs & $\checkmark$ & $\checkmark^{\mathrm{D}}$ & $\checkmark$ & Ranked sequential & NR \\
\midrule
\rowcolor{blue!7}\textbf{Single-iteration study} & FPGA / routed DCP & 16 & $\times$ & $\checkmark^{\mathrm{D}}$ & $\times$ & Single submission & $\checkmark$ \\
\rowcolor{blue!7}\textbf{Multi-iteration study} & FPGA / routed DCP & 16 & $\checkmark$ & $\checkmark^{\mathrm{D}}$ & $\checkmark^{\mathrm{K}}$ & Sequential / branching & $\times$ \\
\bottomrule
\end{tabularx}
\par\vspace{2pt}
\begin{minipage}{\textwidth}
\fontsize{7.3}{8.6}\selectfont
$\checkmark$: explicitly supported; $\times$: excluded by the stated study; NR: not reported in the reviewed version. Timing feedback: candidate implementation measurements guide further optimisation. Physical modification: F, flow re-execution; D, edits to an existing physical design; E, LAMDA's search evaluation replays pre-collected tool measurements. Knowledge reuse: supplementary procedural knowledge or experience across trials, not merely an action catalogue; K, the KB-enabled multi-iteration variant. Control-level study: a common-backend comparison of recipe selection, direct operations and constructed procedures; NR does not exclude other ablations. Designs count backend cases, not models or trials: ORFS evaluates three circuits at two nodes, FluxEDA's entry covers its ECO case and PACT counts DCP instances; both studies here use the same 16 designs.
\end{minipage}
\endgroup
\end{table*}

Existing LLM systems therefore often work around the implemented state rather than improving it directly, generating scripts or templated implementation flows~\cite{chateda}, re-running flow parameters or constraints~\cite{orfs,nexus}, or rewriting RTL through optimisers~\cite{drrtl2026} or end-to-end design agents~\cite{designconductor2026}. Recent ASIC systems move closer to the physical-design problem by editing P\&R designs directly through engineering change orders (ECOs). RSR schedules ECO sizing, threshold-voltage and buffering commands, guided by retrieved expert scheduling strategies and result feedback~\cite{rsr}, while FluxEDA keeps commercial tool sessions alive so that an agent can apply such edits and roll back rejected steps~\cite{fluxeda}. Concurrent with this work, PACT demonstrates LLM-guided ECO tuning of routed FPGA checkpoints by ranking typed edits, applying them sequentially and retaining only valid timing-improving results~\cite{pact}. PACT thereby establishes routed FPGA checkpoint editing as a strong application of LLMs to backend EDA. Yet how LLMs fit into the backend flow remains an open question: how much action freedom can an agent use reliably, how efficiently can it navigate the timing-optimisation search space, and which aspects of its harness best supply backend design expertise. Table~\ref{tab:prior_work} summarises how prior ML- and LLM-based systems address these dimensions.

This paper therefore presents \textsc{LATCH} (LLM-Assisted Timing Closure Harness), a two-layer framework that improves post-route FPGA $F_{\max}$ to investigate how LLMs behave in backend EDA. The deterministic execution layer characterises each checkpoint, applies physical modifications and validates the result. The LLM orchestration layer interprets backend evidence, requests additional analysis when needed and ranks the next action to try. This separation lets the paper study LLM decision-making whilst keeping implementation, validation and measurement deterministic. Contributions are as follows:

\begin{enumerate}
\item We introduce \textsc{LATCH}, a reproducible framework for LLM-orchestrated post-route FPGA $F_{\max}$ improvement. Its protocol-mediated interface isolates LLM reasoning from deterministic EDA execution, providing an extensible framework: \textit{\url{https://github.com/aprilaihub/LATCH}}
\item We contribute nine engineer-written expert optimisation recipes encoding physical-design expertise as executable actions. Each targets a known timing failure mode and is profiled by measured value and validation risk.
\item We characterise the trade-off between FPGA backend action freedom, execution reliability and timing improvement in a single-iteration study that compares three granularities of agent control. We then fix the most reliable recipe tier in a multi-iteration ablation study to explore design search mechanism effectiveness by comparing branching exploration and knowledge base guidance with sequential reasoning and a fitted heuristic.
\item We evaluate the framework on sixteen post-route FPGA designs, using seven LLMs in the single-iteration study, three LLMs in the multi-iteration ablation study, structural legality checks on every candidate, and simulation-based functional checking on the published result.
\end{enumerate}

The single-iteration study identifies a trade-off between action reliability and timing gain: recipes are more reliable, while passing direct operations achieve a slightly higher median improvement. The multi-iteration study therefore fixes the recipe interface to examine how search and knowledge reuse affect optimisation. Branching and knowledge base guidance improve mean gains over sequential reasoning, while comparisons with the fitted heuristic reveal differences in generalisation to held-out designs. The remainder of the paper is organised as follows: Section~\ref{sec:methodology} describes the LATCH framework and expert recipes, Section~\ref{sec:results} presents the benchmarks and both studies, Section~\ref{sec:discussion} discusses limitations and how each can be resolved, and Section~\ref{sec:conclusion} concludes.

\section{Methodology}
\label{sec:methodology}
\begin{figure*}[!t]
    \centering
    \includegraphics[width=0.90\textwidth]{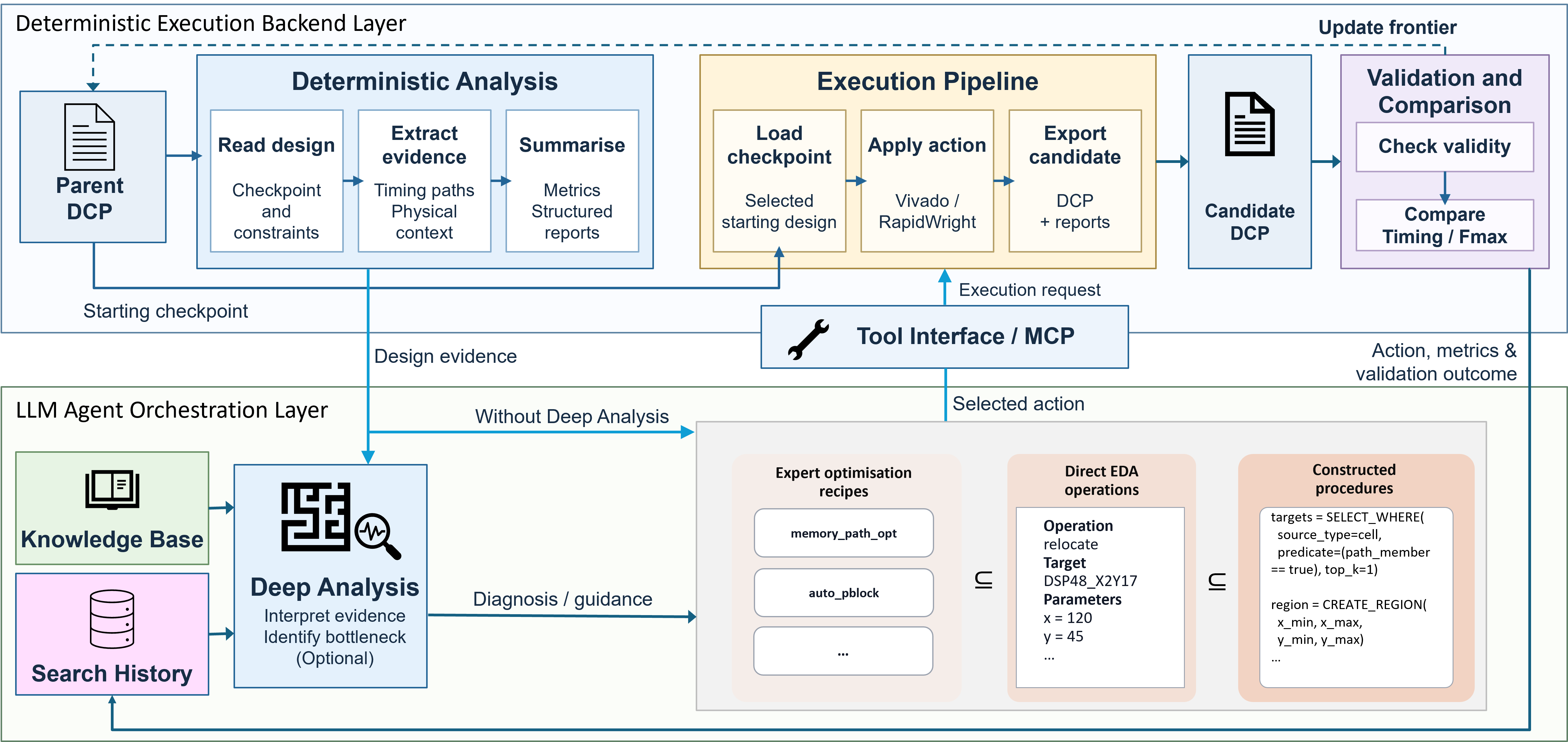}
    \caption{Checkpoint-based FPGA optimisation framework. Design evidence informs the choice of an expert optimisation recipe, direct EDA operations or a constructed procedure. The backend executes proposed modifications and evaluates checkpoints, only retaining validated candidates that improve $F_{\max}$.}
    \label{fig:framework_pipeline}
\end{figure*}

\providecommand{\PipelineFig}{Fig.~\ref{fig:framework_pipeline}}
\providecommand{\CheckpointFig}{Fig.~\ref{fig:checkpoint_search}}

\textsc{LATCH} guides the selection of timing-optimisation procedures for an implemented FPGA design using two vendor tools. AMD Vivado 2025.1 targets FPGA implementation, physical optimisation and timing analysis, and RapidWright 2025.2.1-beta reads and writes the post-P\&R netlist directly to reach transformations Vivado does not expose~\cite{rapidwright}. Both act on a design checkpoint (DCP), which stores the FPGA implemented netlist together with its P\&R. The framework is organised in two layers (\PipelineFig{}): the \textit{deterministic execution backend} performs analysis, modification and measurement, and the \textit{orchestration layer} decides which modification to attempt on a selected DCP, using an agent that reasons over the evidence the backend supplies. The two layers run as servers communicating through the Model Context Protocol (MCP).

\subsection{System Overview}
\label{sec:system_overview}

\subsubsection{Inputs and outputs}

Each optimisation step starts from a retained DCP selected as its parent. The pipeline then produces one candidate DCP from that parent, which joins the retained set if it passes the validation checks of Section~\ref{sec:validation}. The framework therefore improves frequency by selecting among retained DCPs across optimisation steps rather than rebuilding the design from RTL. The run also carries two supporting state objects. The \textit{search history} records the retained checkpoints, their eligibility for expansion, and the ordered actions and measured outcomes leading to each branch, including attempts recorded elsewhere in the search. The \textit{knowledge base} (Section~\ref{sec:knowledgebase}) holds design-independent notes on each recipe, distilled from exhaustive offline trials, and supplies prior expectation where the current evidence is silent.

\subsubsection{The optimisation step}

The optimisation loop starts with Deterministic Analysis, which extracts a fixed evidence set from the parent DCP, including timing reports, critical logic paths, placement spread, and routing detour (defined in Section~\ref{sec:backend_analysis}). This evidence is then provided to the agent in the orchestration layer. When the fixed summary is not sufficient, the agent may request additional read-only probes of the DCP through Deep Analysis to locate the root cause of the timing limitation more precisely. Probes query the cached analysis of the parent, so repeated queries cost no additional tool time, or run additional tool analysis to supply new evidence.

\label{sec:control_interfaces}
When the agent believes it has enough evidence, it performs Action Selection, choosing the next optimisation from the supplied evidence at one of the three nested control levels in Fig.~\ref{fig:framework_pipeline}. At the coarsest, the agent may select from a fixed catalogue of \textit{expert optimisation recipes}, each a sequence of engineer-written transforms targeting a specific timing failure mode. The recipes' internal command sequence and targeting rules are supplied entirely by the framework (Section~\ref{sec:deterministic_backend}). With \textit{direct EDA operations}, the agent gains finer control, specifying a single placement, routing or physical-optimisation call and naming the cells, nets or regions to act on through allow-listed tool command language (TCL) commands; the recipes remain available alongside them. At the finest level, \textit{constructed optimisation procedures} add target construction, allowing the agent to build bounded intermediate targeting, grouping or spatial logic, for example collecting the nets on the critical path, filtering them by routing detour, and routing the remainder. Because all three levels compile to the same fixed set of Vivado and RapidWright commands, the agent's use of each degree of freedom can be evaluated, with tool abstraction, proposal selection and execution performance isolated as separate effects. The selected action then returns to the backend as a typed call, which the Execution Pipeline applies deterministically in Vivado or RapidWright.

\subsubsection{Validation, comparison, and search update}

Validation and Comparison closes the loop, checking whether the candidate checkpoint is legal, re-measuring timing, and comparing the result with its parent. We define an optimisation step as one pass through this loop, from a parent checkpoint to one evaluated candidate. A valid candidate returns to the search space and may become a later parent; an invalid or regressive candidate remains in the run record but is not expanded further.

\begin{figure*}[!t]
    \centering
    \includegraphics[width=0.90\textwidth]{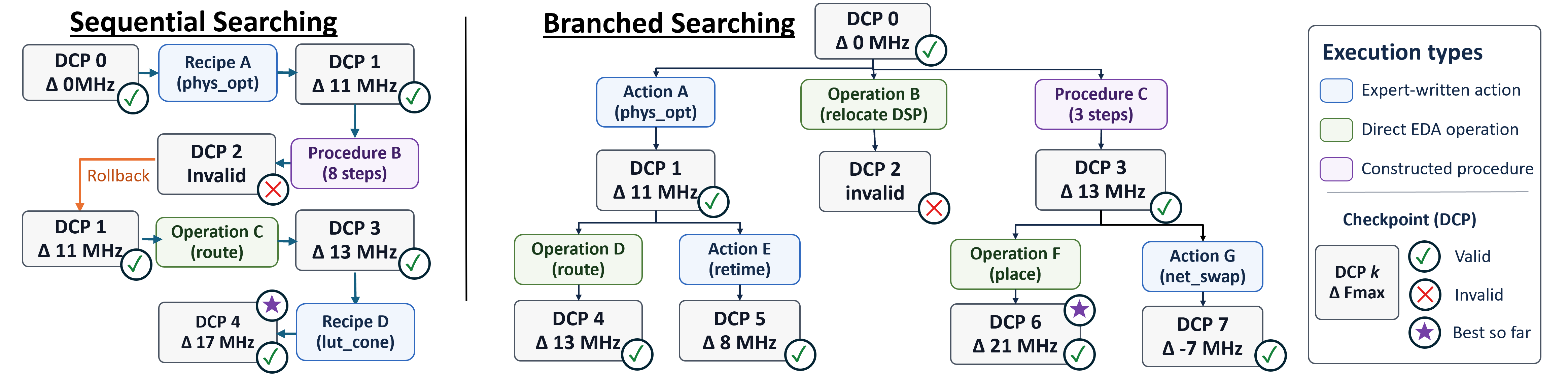}
    \caption{Illustrated example of sequential and branching optimisation patterns with retained checkpoints. An invalid attempt leaves its parent available, and branching allows alternative changes to start from the same checkpoint. The run history records applied changes and measured outcomes. Numerical values illustrate the workflow rather than experimental results.}
    \label{fig:checkpoint_search}
\end{figure*}

\begin{table*}[!t]
\centering
\caption{Expert optimisation recipes. All use Vivado; $\dagger$ additionally uses RapidWright for netlist modification.}
\label{tab:optimisation_flows}
\begingroup
\footnotesize
\setlength{\tabcolsep}{3pt}
\renewcommand{\arraystretch}{1.02}
\begin{tabularx}{\textwidth}{@{}>{\raggedright\arraybackslash}p{3.5cm}X@{}}
\toprule
\textbf{Recipe} & \textbf{Function} \\
\midrule
\texttt{fanout}$^\dagger$ & Splits timing-critical high-fanout nets by a factor of two to eight to reduce loading and routing delay. \\
\texttt{auto\_pblock} & Defines a region from critical-path spread and fabric resources, then unplaces and re-implements the design inside. \\
\texttt{net\_swap} & Targets problematic critical connections to reduce routing-related delay. \\
\texttt{phys\_opt} & Runs Vivado physical optimisation after placement and routing, including replication, cell resizing and register movement. \\
\texttt{lut\_cone}$^\dagger$ & Merges LUT-dominated critical logic cones to reduce logic depth and path delay. \\
\texttt{retime} & Moves registers across combinational logic to reduce delay between sequential stages. \\
\texttt{memory\_path\_opt} & Optimises critical paths involving embedded BRAMs using memory-aware physical optimisation. \\
\texttt{hardblock\_relocate} & Improves physical locality between timing-critical hard blocks and the surrounding logic. \\
\texttt{congestion\_spread} & Redistributes logic in congested regions to reduce routing pressure. \\
\bottomrule
\end{tabularx}
\endgroup
\end{table*}

A run is the ordered set of optimisation steps explored from the initial checkpoint, organised by the two search patterns of Fig.~\ref{fig:checkpoint_search}: sequential and branching. In sequential search, the framework expands one retained checkpoint at a time; in branching search, it keeps multiple retained checkpoints on the frontier and may expand more than one of them within the same run budget. To choose the next expansion, the scheduler ranks legal checkpoint--action pairs across the whole frontier by the priority score the agent assigned to each pair, so the next parent is the checkpoint named by the highest-ranked pair. Branches use the same step mechanism and, acting on different checkpoints, may run concurrently. The run's budget, the number of checkpoints a run may produce, is fixed in advance, as are the depth, expansion and concurrency of branching. Every proposal, action and measured outcome is written to the run record, so a run can be inspected, resumed or reproduced, and the record itself becomes evidence.

\subsection{The Deterministic Execution Backend}
\label{sec:deterministic_backend}

The deterministic backend provides the implementation and evaluation interface between optimisation decisions and the physical FPGA design. The agent may propose backend operations, but analysis probing, execution, checkpoint generation and timing evaluation remain backend-controlled, so optimisation performance is determined by the commands that change the DCP's physical implementation. This section describes how the backend extracts analysis evidence and executes the expert-written recipes, the one action tier the framework authors in advance, each a fixed procedure embedding an engineer's expertise. The finer tiers (Section~\ref{sec:control_interfaces}) have no such fixed procedure, because the agent generates them at run time from the same EDA tool operations.

\subsubsection{Design analysis and diagnostic extraction}
\label{sec:backend_analysis}

Before optimisation, the backend performs a deterministic analysis of the input implementation: Vivado opens the input checkpoint, parses its timing summary to extract the worst negative slack (WNS), total negative slack (TNS), and number of failing endpoints, and queries the clock period to determine the target clock frequency and the achievable operating frequency.

For a clock period $T_{\mathrm{clk}}$ and measured WNS, the backend estimates the achievable frequency as
\begin{equation}
F_{\max} =
\frac{1000}
{T_{\mathrm{clk}}-WNS}
\quad \text{MHz},
\label{eq:fmax}
\end{equation}
where timing violations correspond to $WNS<0$.

Structural and physical diagnostics complement the timing report. Vivado identifies critical high-fanout nets and extracts the cells and pins lying on the most critical paths, which RapidWright uses to measure the physical spread of critical-path cells (maximum and average cell distances) and the routing detour of critical connections, identifying those whose routed length considerably exceeds their direct physical separation. High spread indicates excessive interconnect distance and motivates region-constrained placement, whereas large detours identify candidates for cell re-placement. The resulting diagnostic set $\mathcal{D}$ comprises WNS, TNS, $F_{\max}$, the number of failing endpoints $N_{\mathrm{fail}}$, critical high-fanout nets $\mathcal{N}_{\mathrm{fanout}}$, critical-path information $\mathcal{P}_{\mathrm{crit}}$ and physical placement characteristics $\mathcal{S}_{\mathrm{place}}$.

\subsubsection{Expert optimisation recipes}
\label{sec:backend_optimisers}

The backend exposes a catalogue of expert optimisation recipes, each with a fixed sequence of examination, modification, implementation and evaluation operations directed at one class of timing bottleneck; Table~\ref{tab:optimisation_flows} lists the nine forming the \textit{expert-written recipe} action tier used throughout this study. Because a recipe's internal command sequence is supplied by the framework, selecting it commits the agent to a known implementation procedure rather than an arbitrary tool invocation.

The catalogue is organised around the physical causes of critical-path delay identified by the deterministic and deep analysis. Placement-oriented recipes act on where logic sits: \texttt{auto\_pblock} re-implements the design within a region derived from critical-path spread, suiting designs whose critical logic is dispersed across the fabric, while \texttt{hardblock\_relocate} improves locality between timing-critical hard blocks and the surrounding soft logic. Because embedded block random access memories (BRAMs) and digital signal processing (DSP) columns are fixed in the fabric, \texttt{hardblock\_relocate} acts on the placement and registering of the logic around them, which bounds its yield on memory-dense designs. Routing-oriented recipes act on how logic is connected (\texttt{net\_swap}, \texttt{congestion\_spread} and \texttt{fanout}), with the \texttt{fanout} split factor bounded between two and eight to avoid excessive replication. Logic and sequential recipes act on the structure of the path itself: \texttt{lut\_cone} merges lookup table (LUT)-dominated critical cones to reduce logic depth, and \texttt{retime} moves registers across combinational logic to balance delay between sequential stages. Finally, \texttt{memory\_path\_opt} targets paths through BRAMs, and \texttt{phys\_opt} applies Vivado's general physical optimisation after placement and after routing.

The recipes also differ in where they execute, which matters more for reliability than the ambition of the transformation. Seven perform their modification inside Vivado and inherit its incremental optimisation semantics, under which a transformation is retained only where the tool measures an improvement. The remaining two, \texttt{fanout} and \texttt{lut\_cone}, modify the netlist in RapidWright and write an intermediate checkpoint that Vivado reopens and reroutes, gaining direct access to placement and connectivity that Vivado does not expose but leaving the vendor tool's gated loop: the returned design is checked for legality when reopened, but no intermediate improvement guarantee applies to the modification itself. Section~\ref{sec:ablation_study} shows that validation risk concentrates almost entirely in these two recipes.

\subsubsection{Timing verification and quantitative feedback}
\label{sec:backend_verification}

Every optimisation is evaluated using implementation-level timing data: the backend records the initial WNS and clock period before modification, re-evaluates timing after the candidate has been routed, converts the result to an achievable $F_{\max}$ using \eqref{eq:fmax}, and compares it with the baseline. For an initial implementation $D_0$ and an optimised implementation $D_i$, the changes $\Delta WNS = WNS(D_i)-WNS(D_0)$ and $\Delta F_{\max} = F_{\max}(D_i)-F_{\max}(D_0)$ are positive for an improvement. The backend tracks the best observed WNS across a run, retains TNS and the failing-endpoint count from the initial report, and inspects route status after each physical modification so that routing errors are identified before a timing result is accepted. The objective continues to reward improvement after timing is met. $F_{\max}$ in \eqref{eq:fmax} rises with positive WNS, and no run terminates on reaching $WNS \geq 0$.

\begin{figure}[!t]
    \centering
    \includegraphics[width=0.85\columnwidth]{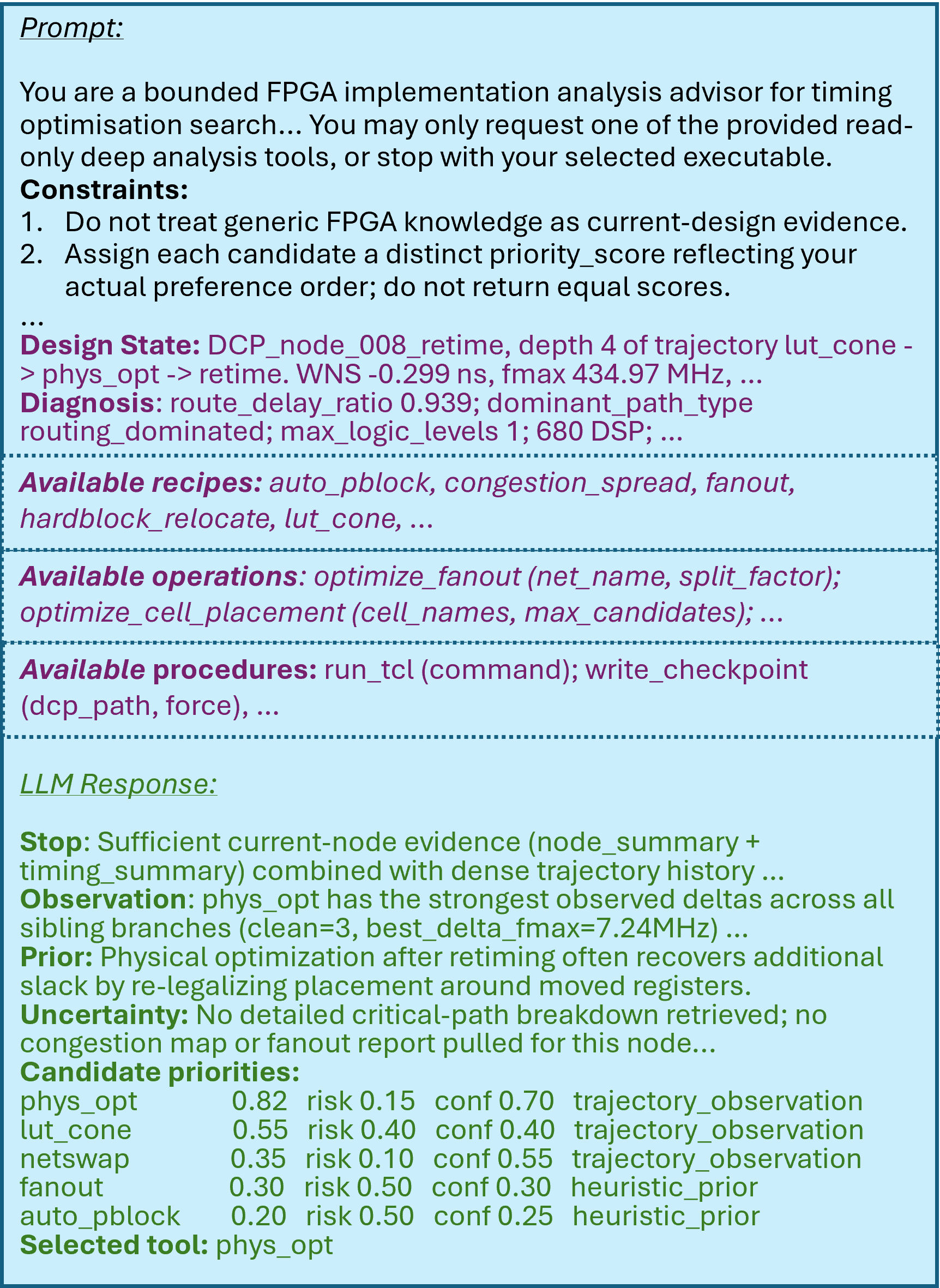}
    \caption{Analysis prompt for a recipe-selection decision. The design state, diagnosis, and available agent control interfaces -- which change depending on which control level is available to the agent -- are injected (\textit{\textcolor{purple}{purple}}), while the underlying template remains (\textit{\textcolor{black}{black}}). A typical agent response is shown (\textit{\textcolor{green}{green}}), ending in the ranked candidate list and the selected action.}
    \label{fig:prompt_listing}
\end{figure}

\begin{figure}[!t]
    \centering
    \includegraphics[width=0.85\columnwidth]{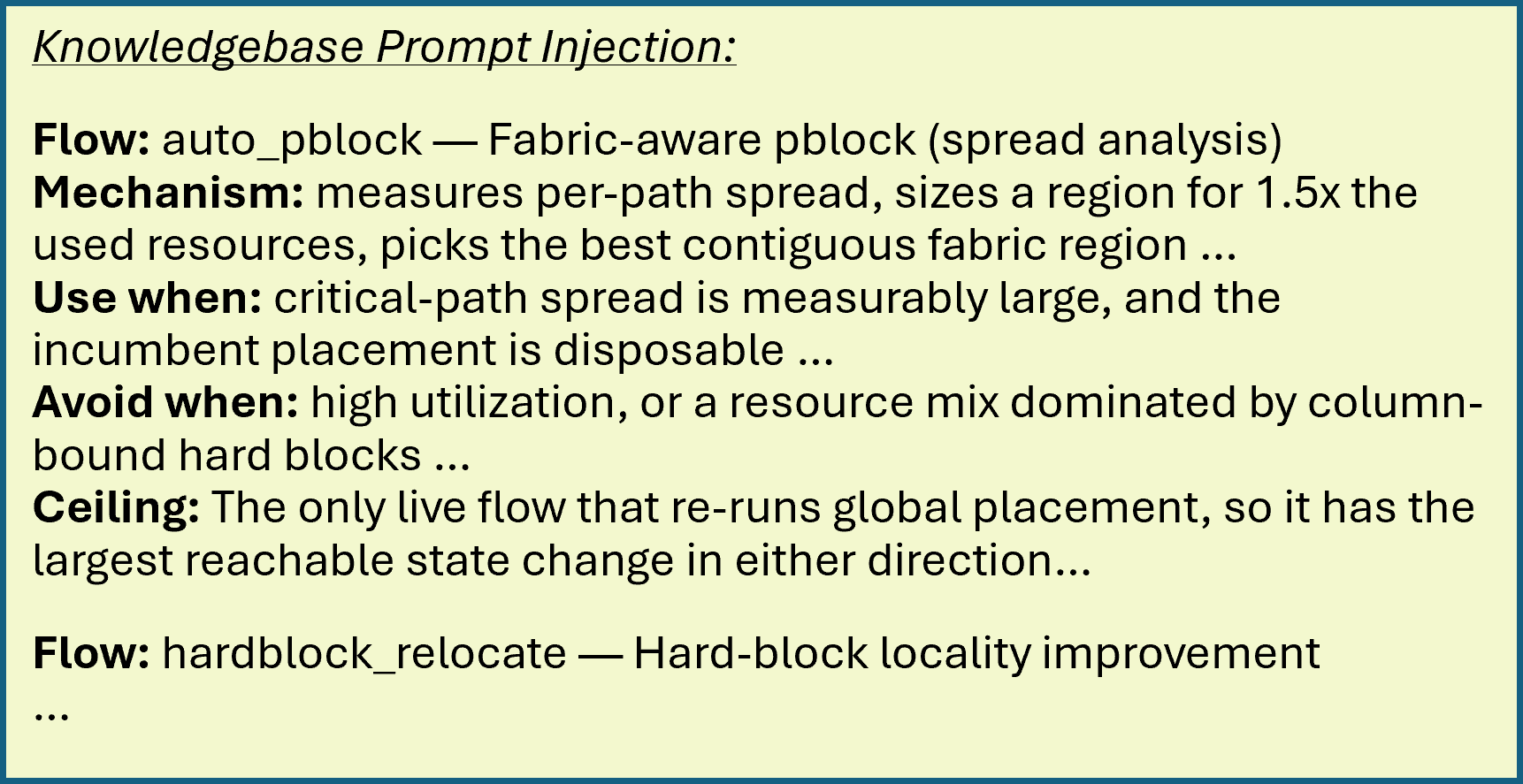}
    \caption{Example knowledge base recipe description, allowing the agent to reason about a recipe's use case from generalised, non-benchmark-specific past information.}
    \label{fig:knowledgebase_listing}
\end{figure}

\providecommand{\ListingOne}{Fig.~\ref{fig:prompt_listing}}
\providecommand{\ListingTwo}{Fig.~\ref{fig:knowledgebase_listing}}

\subsection{The LLM Agent Orchestration Layer}
\label{sec:agent_pipeline}

At each step, the agent occupying the pipeline's decision stage is given the current design state and the actions available at the active control level of Section~\ref{sec:control_interfaces}. What the agent returns therefore depends on this tool abstraction level: at the recipe interface it returns a ranked choice over the optimisation recipe catalogue, whereas at the direct-operation and constructed-procedure interfaces it submits an executable proposal, which it may revise in response to construction feedback before committing.

\subsubsection{Prompt construction}

The diagnostic evidence extracted before each decision (timing summaries, critical-path composition, and the relevant cell, net, and placement context), together with the recipes or operations available at the active control level, is inserted into a fixed prompt template alongside the task description and the required response format (\ListingOne{}). At the recipe interface, the agent returns a ranked list of candidate actions each scored for priority and risk, with a stated confidence and evidence basis. A fixed, deterministic rule then selects the candidate to execute: highest priority wins, with ties broken by the stated confidence, then by lower risk, then by checkpoint and catalogue order, after two filters exclude the recipe last applied to a checkpoint and any recipe sequence already attempted. Stating evidence and uncertainty before committing to a score gives the agent room to reason explicitly and leaves a traceable record of why a recipe was preferred, whether or not that reasoning proves right.

\subsubsection{The knowledge base}
\label{sec:knowledgebase}

This supports action selection at the recipe tier by providing additional information to the agent. Each recipe holds one entry (\ListingTwo{}), distilled from an offline, exhaustive execution of all nine recipes across the known benchmark designs rather than authored from first principles, with the same fields: the mechanism the recipe acts through, the design conditions under which it tends to help or hurt, and the ceiling on what it can achieve. Because checkpoints differ in function, resource mix, and implementation method, numeric thresholds measured on one rarely transfer to another, so findings are generalised into qualitative conditions rather than measured values, for example a large, disposable placement rather than a percentage, so that the same entry describes a recipe's behaviour on a design the exploration never saw.

Entries are injected as an explicitly labelled prior, and the agent is instructed to cite them as prior evidence rather than as evidence about the design in front of it. Backend timing-closure knowledge resides largely in vendor-specific interfaces and engineering experience unlikely to appear in training data; the knowledge base therefore supplies it as in-context learning.

\subsubsection{Deep analysis and the bounded decision loop}

When the cause of a violation remains ambiguous, the agent may request evidence beyond the fixed deterministic analysis through a bounded read-only probe loop, in each turn either issuing probes or returning its selection. Probes expose existing evidence without creating new design states, refining the description of the current DCP only. Ten probes are available in three families: three re-run current-design reports for timing, routing, and resource utilisation; five retrieve established search evidence, including the DCP's state and parent's $F_{\mathrm{max}}$ gain, stored diagnosis, ancestor recipe outcomes, outcomes in other branches, and validation errors and warnings; and two summarise broader search history through recurring recipe sequences and comparable prior cases. The available set is decision-dependent: all ten are offered at the root; eight at ordinary checkpoints, omitting routing and utilisation reports already measured; and seven after validation failure, omitting timing.

This deep analysis loop is used only in the multi-iteration study, as the single-iteration study focuses on splitting planning and tool abstraction. To prevent scaling token costs, it is bounded at six commands for the root decision step and four at every other DCP step. Because this bounds evidence-gathering rather than reasoning, a session can reach the bound before returning a ranking; in the reported experiments, such a session terminated without a model ranking and received a fixed ordering over the legal candidate set at neutral priority. The bound applies identically at every level, so comparisons between levels are unaffected, although a substituted choice describes the fallback rather than the agent. A subsequent sweep with a raised bound reduced substitution to below 1\% of sessions and did not improve achieved frequency, confirming that the reported results are not an artefact of the bound.

\subsubsection{Validation and recovery}
\label{sec:validation}

Every candidate DCP must be shown to be both a physically legal design and to agree with the design it replaces on the applied tests before it can be trusted, using two checks that differ in cost and frequency. Structural validation, which establishes \textit{physical validity}, runs on every candidate: the implementation logs and the route-status report are parsed to confirm that the produced checkpoint reopens, that routing completed with no unrouted nets or routing errors, and that timing can be re-measured on the design's target clock rather than on an all-clock summary. Functional checking, whose success is a \textit{functional pass}, is reserved for the single candidate selected as a run's published result, which in the single-iteration study is each trial's committed candidate, since it requires simulation and establishes agreement for the applied tests rather than formal equivalence: a precheck is followed by five deterministic simulation runs of 200 input vectors each, and the candidate's outputs are compared against those of the starting design; a byte-identical candidate inherits its parent's certificate rather than being re-simulated. An action that proves infeasible to execute is retried once automatically; a candidate that fails either check, or regresses relative to its parent, is excluded from further search and the step reverts to its parent checkpoint, while its execution record is retained in the run history.

\section{Results}
\label{sec:results}
\subsection{Benchmark and Experimental Constraints}
\label{sec:benchmarks}

Evaluation uses sixteen post-route implementation DCPs: fifteen drawn from the FPL 2026 AMD Competition benchmarks ~\cite{amd_benchmarks} and one from RapidWright ~\cite{picoblaze_benchmark}. The set spans roughly two orders of magnitude in instance count, from approximately 3k to 325k LUTs and FFs, covers soft-logic-, memory- and arithmetic-dominated resource mixes, and was selected to span the range of achievable timing slack, with baseline frequencies from 406~MHz, where WNS is under 1~ns, down to 48.2~MHz, where WNS is $-19.2$~ns. Several designs therefore yield substantial frequency to almost any intervention, while others are limited by logic depth or by delay internal to a hard block and hence have limited yield to any method examined here.

The corpus is partitioned into three groups: eight \textit{known} designs (the development set), whose characteristics and performance were assessed before full evaluation and were used to develop the knowledge base and fit the deterministic policy described in Section~\ref{sec:ablation_study}; five \textit{re-implementations} from design families represented in that fitting set; and three \textit{out-of-family} designs from families not represented in it, the latter two groups forming the eight held-out designs. This partition separates designs used to establish and tune the method from those reserved to test generalisation. It applies only to the multi-iteration study, whose knowledge base and fitted heuristic were derived from the known designs; the single-iteration study uses no knowledge base and reports all sixteen designs together.

\providecommand{\ControlFigures}{Figures~\ref{fig:dof_overall}--\ref{fig:dof_heterogeneity}}
\providecommand{\OutcomesFig}{Fig.~\ref{fig:dof_overall}}
\providecommand{\OutcomesFigure}{Figure~\ref{fig:dof_overall}}
\providecommand{\ConversionFigure}{Figure~\ref{fig:dof_funnel}}
\providecommand{\DesignFigure}{Figure~\ref{fig:dof_heterogeneity}}

\subsection{Single-Iteration Agent Control Study}
\label{sec:single_iteration_control}

This study examines how agent control affects the reliability and timing gain of a single optimisation step across 16 FPGA designs and seven LLMs: Claude Sonnet 5, DeepSeek v4 Flash and Pro, Gemini 3.6 Flash, and GPT-5.6 Luna, Sol, and Terra. The dataset contains one selected outcome per design--model pair under Expert optimisation recipes, Direct EDA operations, and Constructed procedures, giving 112 trials per condition and 336 in total. The recipe condition permits only recipe selection; the other conditions retain inherited interfaces but explicitly request direct operations or a constructed procedure. We therefore measure outcomes under capability-use instructions, with actual execution of the requested capability assessed separately. No knowledge base is used in this study.
For each design, all conditions share the starting checkpoint, base observations, read-only query definitions, and planning limits: four model turns, eight query items, three construction-feedback rounds, and normally 65,536 input bytes including interface documentation. The LLM may query the design and correct construction errors before committing at most one proposal for execution and scoring, without subsequent revision using measured candidate timing. Each proposal allows up to eight statements, including at most four placement, routing, or physical-optimisation calls.

PicoBlazeArray uses the 2025.1 checkpoint (starting $F_{\max}=350.8$~MHz). Its 21 selected outcomes come from 38 started engineering attempts, with provider changes and dated DeepSeek endpoints (Flash-0731 and Pro-0813). Continuations follow the approved engineering sequence rather than maximum performance. The final constructed-procedure continuations admitted up to 262,144 input bytes; earlier attempts and direct operations admitted 65,536. These are protocol exceptions rather than independent optimisation repeats or a frozen replay. Three tool-deadline timeouts count as unsuccessful trials in the pass-rate denominator; their unmeasured timing changes are not imputed as zero.

A \textit{physically valid result} passes the execution, output, and routing checks. Functional evaluation follows Section~\ref{sec:validation}. A \textit{functional pass} requires all prescribed checks to pass; \textit{output mismatch} denotes differing outputs, and \textit{functional check incomplete} denotes validation-infrastructure errors preventing completion. The end-to-end functional-pass rate is the fraction of each condition's 112 trials achieving a functional pass.

\begin{figure}[!t]
    \centering
    \includegraphics[width=\columnwidth]{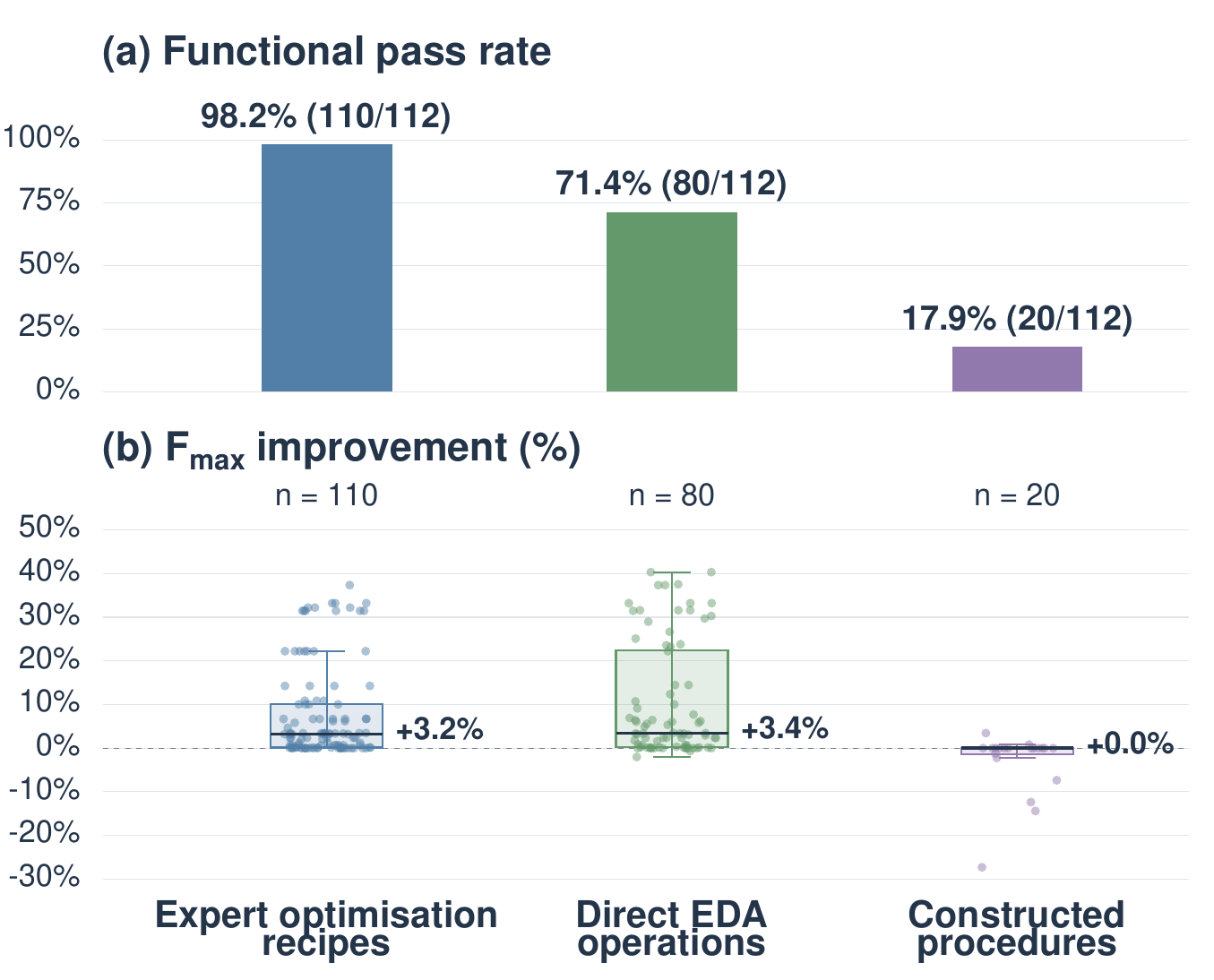}
    \caption{Reliability and timing improvement across 16 designs and seven LLMs (112 trials per condition). (a) End-to-end functional-pass rates, with passing and total trial counts. (b) Percentage $F_{\max}$ improvement relative to each trial's starting checkpoint, for functional-pass trials only. Points show individual trials with horizontal jitter for visibility. Boxes span the interquartile range; whiskers extend to the most extreme observations within 1.5 times that range. Labelled lines mark medians. Zero and negative improvements are retained.}
    \label{fig:dof_overall}
\end{figure}

\begin{figure*}[!t]
    \centering
    \includegraphics[width=\textwidth]{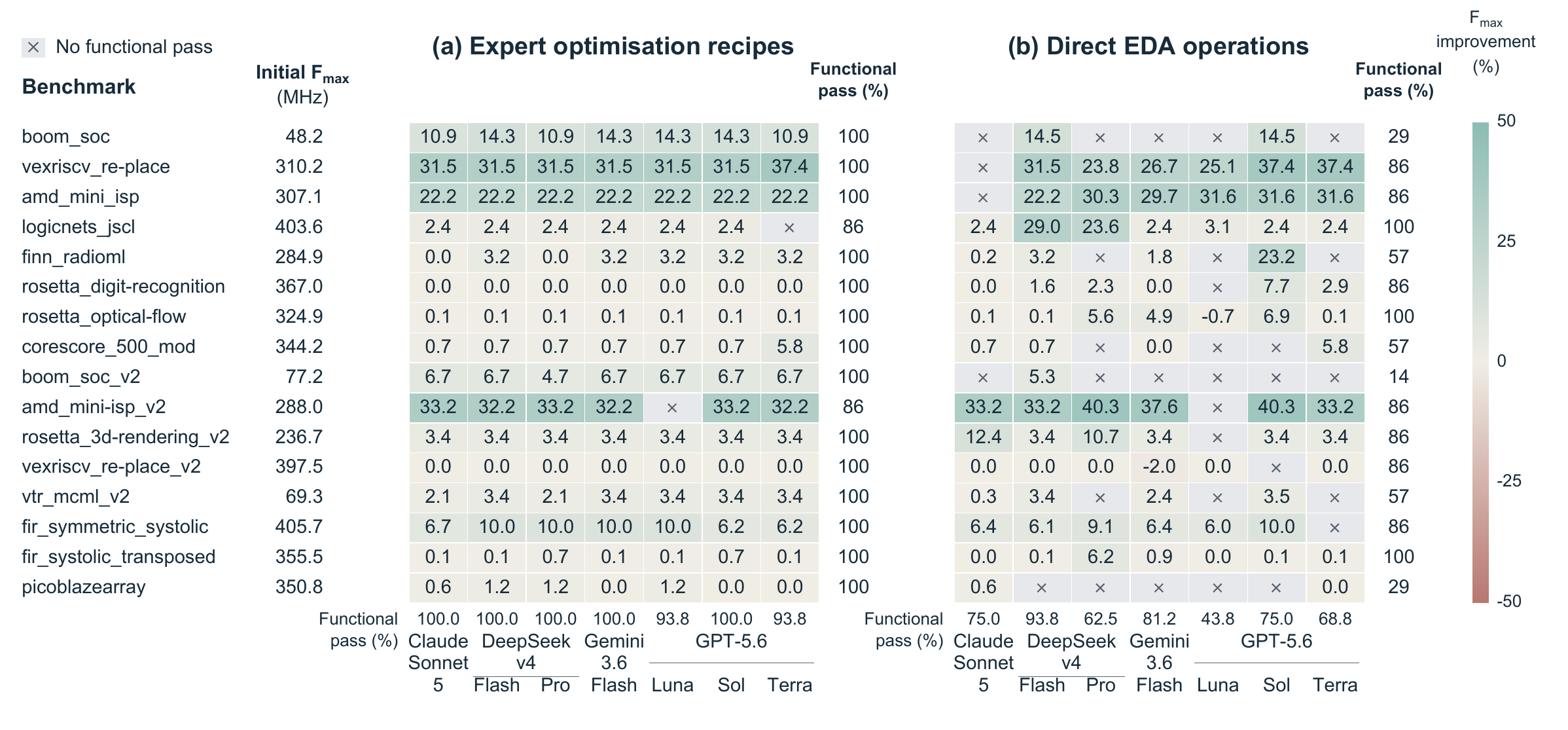}
    \caption{Design--model timing outcomes under (a) Expert optimisation recipes and (b) Direct EDA operations. Each numeric cell shows percentage $F_{\max}$ improvement relative to the starting checkpoint for one functional-pass trial; $\times$ denotes no functional pass. Initial $F_{\max}$ gives the starting frequency in MHz. Designs are ordered by increasing initial $F_{\max}$; both panels use the same design and model ordering and a common colour scale. Row pass percentages are calculated across seven models; column pass percentages are calculated across sixteen designs.}
    \label{fig:dof_heterogeneity}
\end{figure*}

The signed timing change $\Delta F_{\max}$ of Section~\ref{sec:backend_verification} is taken relative to the starting (root) checkpoint, and Figs.~\ref{fig:dof_overall} and~\ref{fig:dof_heterogeneity} report it as per-trial percentage improvement, \(100\,\Delta F_{\max}/F_{\max}^{\mathrm{root}}\); implementation case studies use MHz. Byte-identical checkpoints receive zero change. Distributions and the design--model matrix retain all functional passes. Timing improvement requires a functional pass and \(\Delta F_{\max}>0\); outcome-rate denominators include all assigned trials.

\begin{figure}[!t]
    \centering
    \includegraphics[width=\columnwidth]{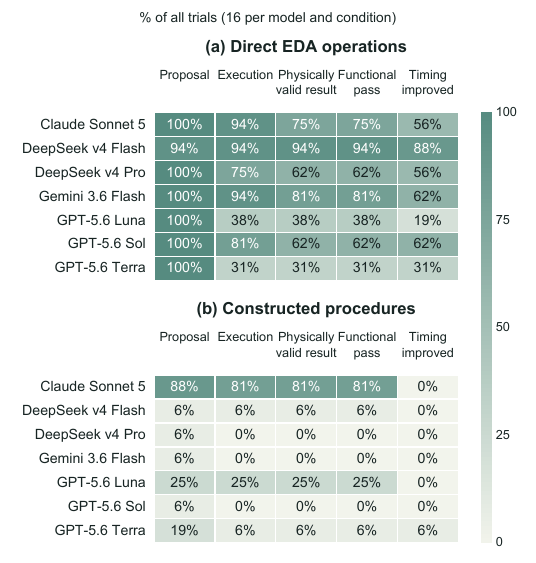}
    \caption{Progression from proposal generation to timing improvement for (a) Direct EDA operations and (b) Constructed procedures. Each row represents 16 trials for one LLM. Percentages are calculated over all 16 trials, with each stage retaining only trials that satisfy all preceding stages. The final stage requires a functional pass and $\Delta F_{\max}>0$.}
    \label{fig:dof_funnel}
\end{figure}

\begin{table*}[!t]
    \centering
    \caption{Physical case studies. Selected and Best denote the LLM-selected recipe and best recorded recipe from the same starting checkpoint; Direct denotes direct operations. Delays describe each output's worst setup path. Resources describe the input.}
    \label{tab:dof_physical_cases}
    \small
    \setlength{\tabcolsep}{3pt}
    \renewcommand{\arraystretch}{0.98}
    \begin{tabular}{@{}llrrrrl@{}}
        \hline
        Circuit / model & Output & $\Delta F_{\max}$ & WNS & Logic & Route & Limiting-path structure \\
         & & (MHz) & (ns) & (ns) & (ns) & \\
        \hline
        LogicNets / DeepSeek v4 Flash & Selected recipe & +9.5 & $-0.921$ & 0.514 & 1.887 & LUT / multiplexer path \\
        \footnotesize 30.8k LUTs, 1.7k FFs, & Best recorded recipe & +98.0 & $-0.494$ & 0.509 & 1.467 & LUT / multiplexer path \\
        \footnotesize 0 DSPs, 0 BRAMs & Direct operations & +117.0 & $-0.421$ & 0.562 & 1.340 & LUT / multiplexer path \\
        \hline
        FINN RadioML / GPT-5.6 Sol & Selected recipe & +9.1 & $-1.801$ & 1.135 & 2.198 & 12-level carry/LUT path \\
        \footnotesize 73.6k LUTs, 45.8k FFs, & Best recorded recipe & +54.4 & $-1.347$ & 1.155 & 1.566 & 12-level carry/LUT path \\
        \footnotesize 0 DSPs, 24.5 BRAMs & Direct operations & +66.1 & $-1.249$ & 2.405 & 0.196 & Six-RAMB36E2 cascade \\
        \hline

    \end{tabular}
\end{table*}

Fig.~\ref{fig:dof_overall} summarises functional-pass rates and timing improvements across the three control conditions. Recipes achieve a 98.2\% pass rate (110/112), direct operations 71.4\% (80/112), and constructed procedures 17.9\% (20/112). All recorded physically valid candidates pass the functional checks; the remaining trials yield no validated functional-pass output. Among functional passes, direct operations yield a slightly higher median improvement than recipes (+3.4\% versus +3.2\%), but improve timing in fewer of the 112 assigned trials (67 versus 91). Constructed procedures have a median improvement of 0\%: 10 of their 20 passing trials show no change, eight regress, and two improve.

Fig.~\ref{fig:dof_funnel} tracks progression from qualifying proposals---executable submissions in the requested interface---to timing improvement. Execution requires a direct EDA operation beyond recipe selection in panel (a), or use of a procedure-created intermediate object by a subsequent operation in panel (b). Unlike Fig.~\ref{fig:dof_overall}, later stages count only trials that also execute the requested capability. Of the 112 direct-operation trials, 111 produce qualifying proposals; 81 (72.3\%) execute the capability, 71 pass functional checks, and 60 improve timing. For constructed procedures, only 25 of 112 trials produce qualifying proposals; 19 execute and pass, and one improves timing, so their largest loss occurs before a qualifying proposal is produced. Model differences are also substantial: DeepSeek v4 Flash executes direct operations and passes on 15 designs, improving 14, whereas Claude Sonnet 5 executes constructed procedures and passes on 13 designs but improves none, so successful execution does not necessarily yield a timing gain.

Fig.~\ref{fig:dof_heterogeneity} shows that the effect of additional control depends on both design and LLM. On \texttt{vexriscv\_re-place}, switching from recipes to direct operations increases the timing improvement from +31.5\% to +37.4\% for GPT-5.6 Sol, but reduces it from +31.5\% to +23.8\% for DeepSeek v4 Pro. On \texttt{boom\_soc}, all seven LLMs pass with recipes, whereas only DeepSeek v4 Flash and GPT-5.6 Sol pass with direct operations, with similar gains to their recipe results. Additional control can therefore increase or reduce timing gains, or fail to produce a functional-pass result.

\begin{figure*}[!t]
    \centering
    \includegraphics[width=0.9\textwidth]{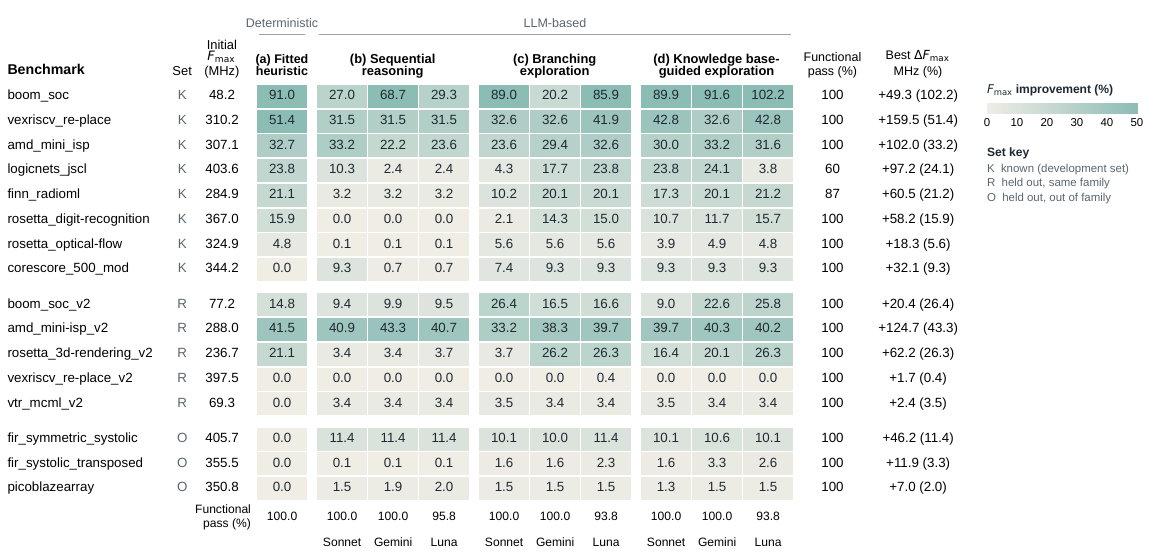}
    \caption{Best functional-pass $F_{\max}$ improvement per benchmark for \textbf{(a)} the deterministic fitted heuristic and the LLM-based levels \textbf{(b)} sequential reasoning, \textbf{(c)} branching exploration and \textbf{(d)} knowledge base-guided exploration, each with Claude Sonnet 5, Gemini 3.6 Flash and GPT-5.6 Luna (best of three repeats).}
    \label{fig:per_benchmark_fmax}
\end{figure*}

\begin{table*}[t]
    \centering
    \caption{Implementation results for three designs, comparing the input checkpoint with the ablation levels.}
    \label{tab:ablation_physical_cases}
    \small
    \setlength{\tabcolsep}{2.5pt}
    \renewcommand{\arraystretch}{0.98}
    \begin{tabular}{@{}l l r r r r l@{}}
        \hline
        \shortstack[l]{Circuit / Model /\\Resources} & Output & \shortstack{$\Delta F_{\max}$\\(MHz)} & \shortstack{WNS\\(ns)} & \shortstack{Logic\\(ns)} & \shortstack{Route\\(ns)} & Limiting-path structure \\
        \hline
        \multirow{5}{*}{
            \begin{tabular}{@{}l@{}}
                BOOM SoC\\
                \footnotesize GPT-5.6 Luna\\
                \footnotesize 227k LUTs, 98k FFs,\\
                \footnotesize 61 DSPs, 161 BRAMs
            \end{tabular}
        }
        & Input checkpoint & -- & $-19.162$ & 2.001 & 18.471 & 27-level LUT path to load queue \\
        & Fitted heuristic & +43.9 & $-9.283$ & 2.072 & 8.675 & 20-level LUT path to I-cache BRAM \\
        & Sequential reasoning & +14.1 & $-14.463$ & 1.220 & 15.001 & 16-level LUT path to register file \\
        & Branching exploration & +41.4 & $-9.583$ & 2.623 & 8.351 & 34-level LUT/multiplexer path to register file \\
        & Knowledge base-guided & +49.3 & $-8.685$ & 1.982 & 8.314 & 19-level LUT/carry path to register file \\
        \hline
        \multirow{5}{*}{
            \begin{tabular}{@{}l@{}}
                AMD mini-ISP v2\\
                \footnotesize Gemini 3.6 Flash\\
                \footnotesize 3.2k LUTs, 4.5k FFs,\\
                \footnotesize 40 DSPs, 12 BRAMs
            \end{tabular}
        }
        & Input checkpoint & -- & $-1.902$ & 0.394 & 2.711 & 5-level LUT path to line-buffer BRAM \\
        & Fitted heuristic & +119.5 & $-0.884$ & 1.909 & 0.397 & 6-level DSP multiply path \\
        & Sequential reasoning & +124.7 & $-0.853$ & 1.923 & 0.350 & 6-level DSP multiply path \\
        & Branching exploration & +110.2 & $-0.941$ & 1.911 & 0.451 & 6-level DSP multiply path \\
        & Knowledge base-guided & +116.2 & $-0.904$ & 1.925 & 0.400 & 6-level DSP multiply path \\
        \hline
        \multirow{5}{*}{
            \begin{tabular}{@{}l@{}}
                Symmetric FIR\\
                \footnotesize Claude Sonnet 5\\
                \footnotesize 0.7k LUTs, 22.6k FFs,\\
                \footnotesize 680 DSPs, 0 BRAMs
            \end{tabular}
        }
        & Input checkpoint & -- & $-0.465$ & 0.078 & 1.999 & Delay-line register to DSP input \\
        & Fitted heuristic & +0.0 & $-0.465$ & 0.078 & 1.999 & Unchanged; only attempt rejected \\
        & Sequential reasoning & +46.2 & $-0.213$ & 0.995 & 1.048 & 2-level DSP adder-tree cascade \\
        & Branching exploration & +41.1 & $-0.238$ & 0.130 & 1.963 & Valid signal to DSP clock enable \\
        & Knowledge base-guided & +41.1 & $-0.238$ & 0.130 & 1.963 & Valid signal to DSP clock enable \\
        \hline
    \end{tabular}
\end{table*}

Table~\ref{tab:dof_physical_cases} compares selected recipes, the highest-$F_{\max}$ recorded functional-pass recipes from the same starting checkpoints, and direct-operation results, with execution traces and critical-path reports revealing the corresponding physical changes. On LogicNets, DeepSeek v4 Flash unroutes and unplaces the design, then applies placement, physical optimisation, routing, and final physical optimisation; its +117.0 MHz gain exceeds both its selected \texttt{phys\_opt} recipe (+9.5 MHz) and the best recorded recipe (+98.0 MHz), whereas DeepSeek v4 Pro's +95.2 MHz direct-operation gain exceeds only its selected recipe. On FINN RadioML, GPT-5.6 Sol re-places the design and applies physical optimisation and routing, gaining +66.1 MHz as the worst setup path shifts from the 12 carry/LUT levels seen in both recipe outputs to six cascaded RAMB36E2 blocks, changing the structure limiting frequency. On \texttt{vexriscv\_re-place}, DeepSeek v4 Pro's direct operations yield less gain than its selected recipe: lower logic delay is outweighed by increased routing delay (2.160 versus 1.840 ns) on the worst setup path.

\subsection{Multi-Iteration, Expert Recipes Ablation Study}
\label{sec:ablation_study}

Where the single-iteration study asked how much control to give the LLM within one attempt, this study asks how repeated attempts should be organised to efficiently navigate the design space. Given the single-iteration trade-off between reliability and reward, this study fixes the action tier at the nine expert recipes of Table~\ref{tab:optimisation_flows}, which execute identically in both studies, and varies only the selection mechanism over multi-iteration runs to isolate selection from execution. Four mechanisms are compared. The \textit{fitted heuristic} ranks recipes from measured physical features without a language model. \textit{Sequential reasoning} replaces that ranking with LLM-driven selection along a sequential trajectory. \textit{Branching exploration} retains that procedure and adds parallel expansion, evaluating several candidates from one checkpoint and continuing from the best that validates. \textit{Knowledge base-guided exploration} retains branching and adds the knowledge base described in Section~\ref{sec:agent_pipeline}. The last three form a strict inclusion chain, so differences between adjacent levels isolate the added mechanism, while the heuristic is an independent deterministic baseline sharing the same action tier and validation gating. All levels receive the same maximum budget of twelve DCPs per benchmark run. Three of the seven previously characterised LLMs span the cost--capability range: GPT-5.6 Luna and Gemini 3.6 Flash as low-cost options, the latter among the stronger prior performers, and Claude Sonnet 5 as a high-capability reference.

The heuristic is a deterministic, non-learning baseline that establishes the study's floor condition, fitted onto the DCP and recipe results from the known benchmarks .... At each step it measures four properties of the current design state: the proportion of critical-path delay attributable to routing versus logic, the depth of the critical logic chain, the count of instantiated hard-block resources, and the load on the most heavily driven net, each selected for a documented physical link to at least one recipe's mechanism of action and for statistical prominence in differentiating the best recipe in the prior study. It then applies the single highest-scoring recipe; it never branches, hedges between candidates, or revises its scoring rule within a run. For recipe $f$ and standardised feature vector $\mathbf{z}$, the score is
\begin{equation}
\mathrm{score}_f = \beta_f + \sum_i w_{f,i} \, z_i,
\label{eq:heuristic_score}
\end{equation}
where $\beta_f$ is the recipe's mean percentage frequency improvement across the fitting corpus, and the summation is the design-conditioned adjustment; selection is by $\arg\max$ with deterministic tie-breaking. A recipe therefore starts from its average worth and is moved up or down by how far the current design differs from the fitting corpus. Coefficients were fitted by ridge regression against percentage rather than absolute improvement, so that designs with dissimilar baseline clocks contribute comparably, with regularisation strength selected by sweeping several orders of magnitude under leave-one-design-out validation.

Fig.~\ref{fig:per_benchmark_fmax} reports, for every benchmark, the best functional-pass $F_{\max}$ improvement of the fitted heuristic and of each LLM at each LLM-based ablation level. The fitted heuristic and knowledge base-guided exploration reach equivalent corpus means, 19.9\% and 20.3\%, against 10.5\% for sequential reasoning and 16.0\% for branching exploration, but the per-design results show opposite successes. On the known designs, the fitted heuristic shows the strongest $F_{\mathrm{max}}$ improvement, because it was fitted on those circuits, yet on the held-out designs it returns no improvement on five of eight, and on the known \textit{corescore\_500\_mod}. On \textit{picoblazearray}, whose 660 memory blocks lie far beyond the fitted range of 250, the policy selected \texttt{auto\_pblock}, which regressed the design, giving an example where the fitted heuristic cannot reliably extrapolate its selection rule to sufficiently unusual designs. On two further designs, \textit{corescore\_500\_mod} and \textit{fir\_systolic\_transposed}, the selected recipe \texttt{auto\_pblock} could not place the design inside its region, so routing failed on an unplaced netlist, and on the remaining three the selected recipes executed successfully but regressed the design. A fitted policy carries no representation of its own uncertainty, so an out-of-range design produces a confident commitment instead of an abstention, and adding a rejection rule to \eqref{eq:heuristic_score} would drive the heuristic to overfit the corpus, an issue already requiring careful management. In contrast, the three model-driven ablations return a positive gain on 14, 16, and 15 of 16 designs respectively, including every out-of-family design, suggesting an ability to generalise across unseen benchmarks.

Table~\ref{tab:ablation_physical_cases} examines this generalisation at the physical level, taking one design from each partition with the LLM that achieved its best result. On BOOM SoC, GPT-5.6 Luna with knowledge base-guided exploration improves $F_{\max}$ by 49.3~MHz: the framework diagnosed the worst path as routing-dominated, with 18.5 of its 20.5~ns spent in wiring, and applied \texttt{memory\_path\_opt}, whose re-placement reduced that route delay to 8.3~ns. On the held-out AMD mini-ISP, Gemini 3.6 Flash achieves the largest gain in the study, 124.7~MHz; here every level removes the wire-bound path into a line-buffer BRAM, after which the limit lies inside a DSP multiplier, so the levels converge on similar results. On the out-of-family symmetric FIR filter, Claude Sonnet 5 improves $F_{\max}$ by 46.2~MHz where the fitted heuristic returns nothing. In this case, the reason was the valid signal driving the clock enables of the DSP adder tree: a second \texttt{phys\_opt} pass re-optimised that path and moved the limit into the adder tree itself, whereas Sonnet's branching runs stopped after a single pass.

Every ablation nonetheless performs better on designs it has already seen: mean improvement on the development set against the held-out set is 30.1\% against 9.7\% for the fitted heuristic, 12.7\% against 8.4\% for sequential reasoning, 20.9\% against 11.1\% for branching exploration, and 28.6\% against 11.9\% for knowledge base-guided exploration; since the knowledge base was distilled from the same known designs, part of its advantage remains design-specific.

Sequential reasoning underperforms structurally rather than through weaker inference. Approximately 68\% of a design's best observed result is reachable within a single intervention, so a single-trajectory search commits to one opening and cannot recover from a poor one, nor efficiently from an invalid checkpoint, because expanding one candidate at a time forfeits the remaining budget. Evaluating three openings instead of one gives breadth at the root rather than depth and accounts for approximately $+8.6$~MHz of gain, close to two-thirds of branching exploration's 14.2~MHz advantage over sequential reasoning. Breadth pays at the root because depth rewards little beyond it: within branching-capable runs, the share of physically valid DCP improvement attempts achieving no further gain over their own original DCP rises from 9.2\% among the opening candidates to 34.4\% beyond them, so roughly a third of the checkpoint budget beyond the opening step is spent on attempts that do not advance the search. This stems from the recipes themselves: they act only on currently-failing paths and commit only improvements, so repeated application searches already-harvested ground; for example, \texttt{phys\_opt} returns approximately $+22$~MHz on first application and under $+5$~MHz on repetition.

\begin{figure*}[!t]
    \centering
    \includegraphics[width=0.85\textwidth]{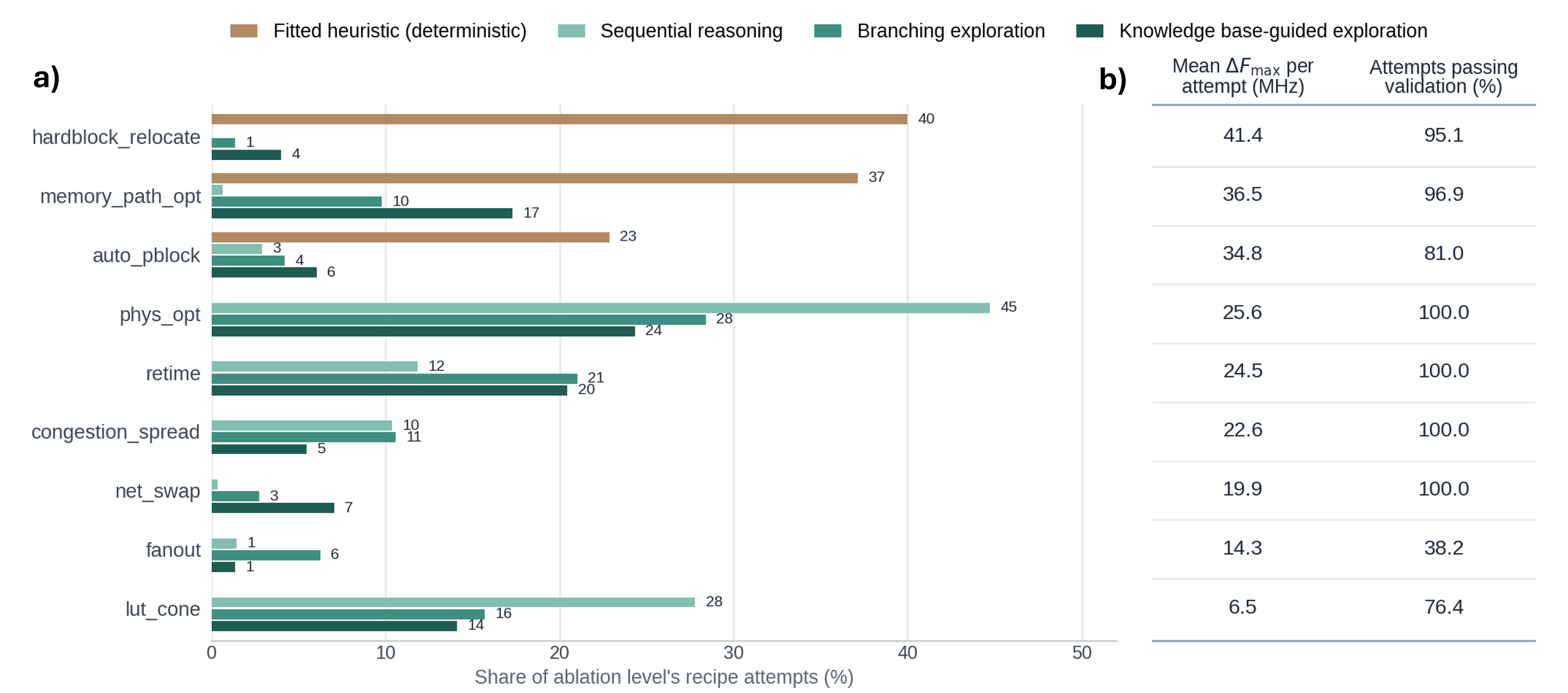}
    \caption{Effectiveness of the expert recipes across the multi-iteration study. \textbf{(a)} Share of each ablation level's recipe attempts spent on each recipe. \textbf{(b)} Each recipe's mean $F_{\max}$ improvement per attempt, measured from the starting DCPs, and the share of its attempts passing physical validation.}
    \label{fig:flow_usage}
\end{figure*}

\begin{figure*}[!t]
    \centering
    \includegraphics[width=0.85\textwidth]{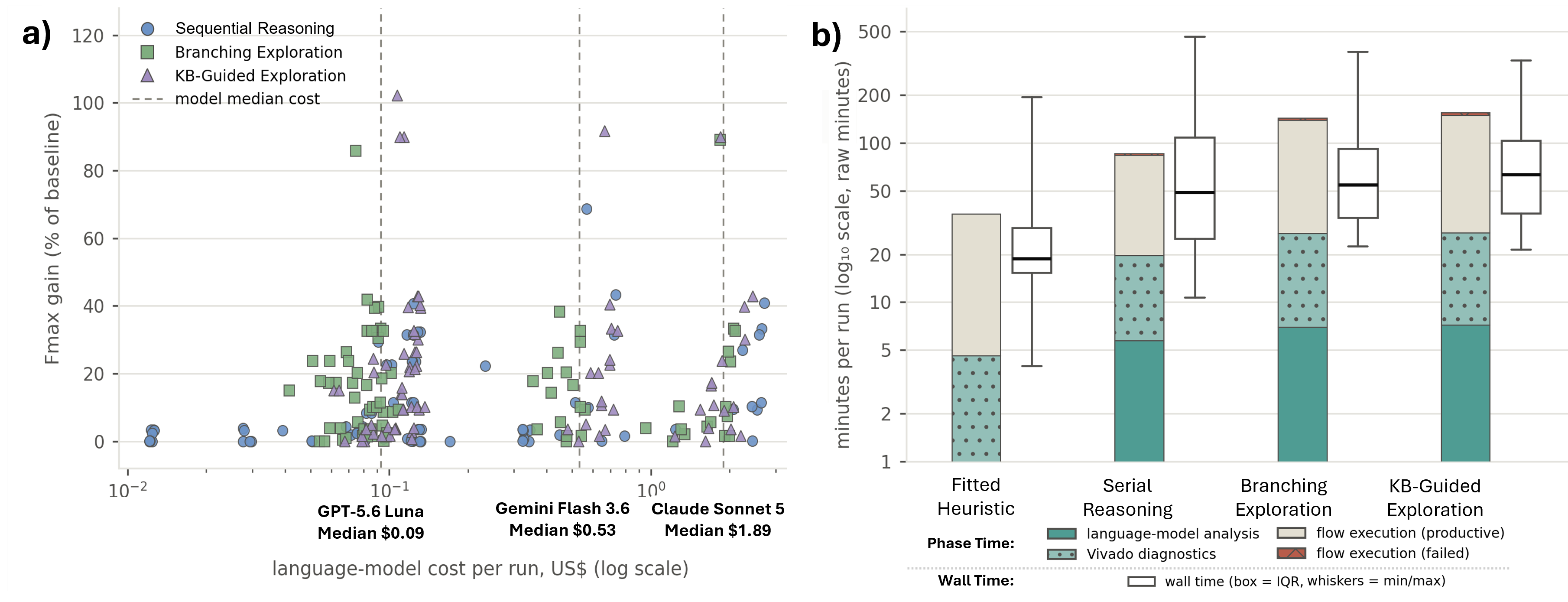}
    \caption{Cost assessment across multi-iteration runs, \textbf{(a)} language-model expense against $F_{\max}$ improvement for every model-driven run, \textbf{(b)} phase time decomposed by pipeline stage beside the elapsed wall time.}
    \label{fig:cost_and_time}
\end{figure*}

Fig.~\ref{fig:flow_usage} relates recipe usage to each recipe's average cumulative $F_{\mathrm{max}}$ improvement over the starting DCP and validation pass rate. The heuristic spends 77\% of its attempts on the two highest-yielding recipes (Fig.~\ref{fig:flow_usage}a), against 0.6\% for sequential reasoning and 11\% for branching exploration, so selection, not execution, is the binding constraint on the model-driven levels; the knowledge base recovers this to 21\% by embedding expert knowledge without altering recipes. Fig.~\ref{fig:flow_usage}b tabulates each recipe's mean gain beside its validation pass rate, and what each recipe physically changes explains the remaining per-design variation. Hard-block-oriented recipes cannot relocate fixed memory and arithmetic columns, so their effectiveness depends on how much surrounding soft logic can be profitably re-placed. This is evident in the BOOM SoC case of Table~\ref{tab:ablation_physical_cases}, where the recipe's outcome varies with the starting placement. Furthermore, \texttt{hardblock\_relocate} is the best-performing recipe on six of sixteen designs, yet, on \textit{picoblazearray}, its 660 small cores already sit beside their memory blocks, leaving little scope for re-placement. Second, \texttt{auto\_pblock} unplaces and re-places the whole design within one region, so its risk lies in whether a given design fits that region rather than in the timing limits: it is reliable on twelve designs, but it fails on over 89\% of attempts on the other four.

Validation outcomes are likewise structural: four recipes, executing wholly within Vivado, passed validation on all 1,477 of their attempts, whereas \textit{fanout} and \textit{lut\_cone}, the two recipes that export the netlist to RapidWright and write it back, forfeiting that guarantee, passed on only 38.2\% and 76.4\%. Validation critically guarded against results that only appear to succeed: on \textit{fir\_systolic\_transposed}, a partially placed \textit{auto\_pblock} output reported a 105~MHz gain with positive slack and was rejected. In this study, correctness was therefore secured by the execution boundary rather than by the reasoning that selected the action.

\begin{figure*}[!t]
    \centering
    \includegraphics[width=0.85\textwidth]{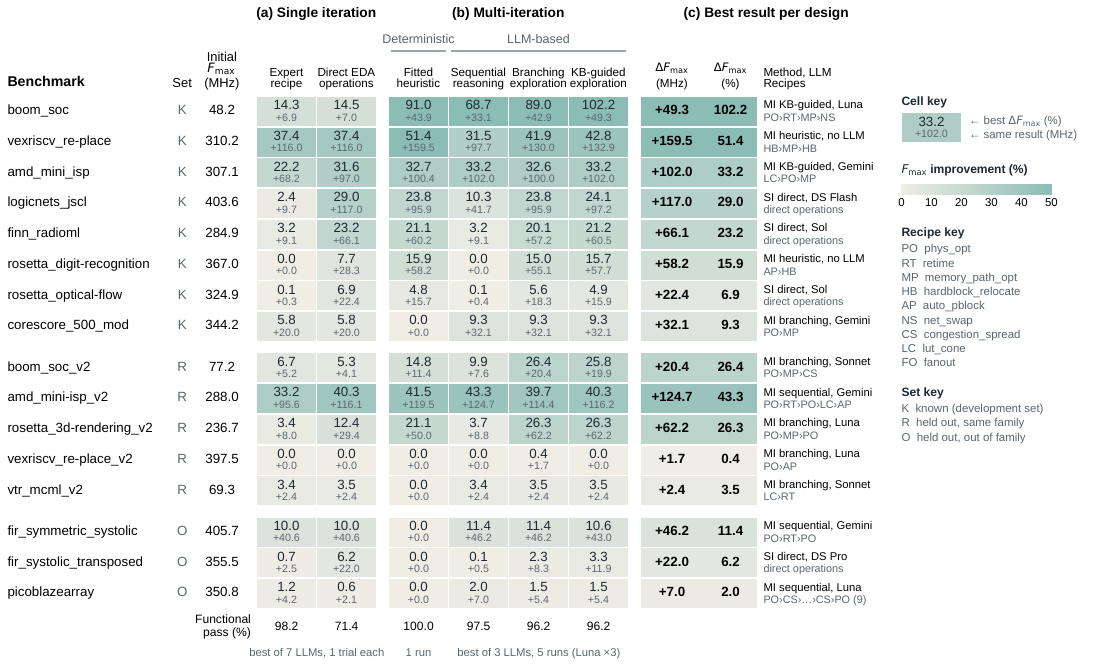}
    \caption{Summary of both studies: best functional-pass $F_{\max}$ improvement per benchmark. \textbf{(a)} Single-iteration study: best of seven LLMs, one trial each, under expert recipes and direct EDA operations. \textbf{(b)} Multi-iteration study: the deterministic fitted heuristic (one run) and the LLM-based levels (best of three LLMs over five runs, three of them GPT-5.6 Luna repeats). \textbf{(c)} The best result per design, with the method, LLM and recipe sequence that achieved it.}
    \label{fig:summary}
\end{figure*}

Two measures of duration assess the cost of running the optimisation (Fig.~\ref{fig:cost_and_time}). \textit{Phase time} attributes each attempt's duration to the pipeline stage responsible and sums across attempts; \textit{wall time} is the run's own elapsed duration. Their ratio indicates the degree of concurrency achieved: it averages 1.66$\times$ for branching exploration and 1.69$\times$ for knowledge base-guided exploration (Fig.~\ref{fig:cost_and_time}b), below the ideal 3.0$\times$ under full overlap of three branch candidates, while remaining below unity for non-branching levels because some elapsed time is not attributed to a labelled stage. This suggests strong utilisation but opportunity for optimising runtime further. Median language-model cost per run varies roughly twentyfold across the three LLMs for a difference in mean improvement of under a percentage point (Fig.~\ref{fig:cost_and_time}a), so among the three LLMs tested, model choice mainly affected cost rather than achieved frequency, consistent with the single-iteration study, in which no LLM dominated across designs. Language-model analysis accounts for 4.6--6.7\% of phase time and tool execution with diagnostics for 91--92\% at every level, so the economic case for orchestration rests on issuing effective tool invocations, not on inexpensive inference. The fitted heuristic finishes sooner only because its ranking is fixed before the run begins: once its top-ranked recipes are spent, it self-terminates after two to four checkpoints, where the model-driven levels re-diagnose and consume the full allowance. Per attempt it is the slowest level in the study, because its two highest-scoring recipes (\texttt{hardblock\_relocate} and \texttt{memory\_path\_opt}) are also the longest to run.

\section{Discussion}
\label{sec:discussion}

Comparing the studies design by design exposes effects hidden by their corpus means (Fig.~\ref{fig:summary}). In the single-iteration study, direct operations matched or exceeded the best recipe result on fourteen of the sixteen designs, despite passing functional checks in 71.4\% of trials against 98.2\% for recipes. The multi-iteration study reverses this pattern: iterative recipe selection exceeds the best single recipe attempt on all sixteen designs and exceeds the best direct-operation attempt on eleven, with one tie. The four remaining direct-operation wins are narrow, ranging from 1.3 to 4.9 percentage points. This contrast is consistent with approximately 68\% of a design's best observed result being reachable in one intervention: direct operations can access an effective intervention immediately, whereas recipes gain from selecting and composing reliable actions. An effective deployment could therefore begin with recipes and escalate to direct operations when a measured bottleneck remains unresolved. Further study should establish which design evidence could trigger this change in agent control and how the remaining budget should be divided between action breadth and continued recipe search.

Unlike the fitted heuristic, the LLM-based harness improves unseen designs because physical bottlenecks vary too widely for a fixed selection rule to remain well calibrated. Among the LLMs tested, model choice changed cost by roughly twentyfold in the multi-iteration study but mean improvement by under a percentage point, and no LLM dominated across designs in the single-iteration study. This result places the main opportunity in the harness rather than in model selection alone: broader recipes, better evidence retrieval, and explicit uncertainty handling could improve coverage and generalisation. The present catalogue remains weighted toward placement and routing, with only \textit{lut\_cone} restructuring combinational logic. Critical-cone technology remapping, don't-care-based resynthesis, or deeper pipelining could reach logic-dominated bottlenecks, but netlist-rewriting actions would require formal or broader equivalence checking beyond the sampled simulation used here. Structural and functional validity also does not ensure a frequency gain when an action crosses a tool boundary, so a commit-only-if-improved check should be reapplied at each boundary.

The harness evidence remains bounded by one device family, toolchain, and tool version, and by a knowledge base distilled from the eight known designs. Larger benchmark sets, other vendors, and independently constructed or continuously updated knowledge bases are needed to test whether the harness is truly transferable. Finally, the same checkpoint-and-validate architecture could target power and area improvement, at the added complexity of multi-objective trade-offs, and its MCP-based interface provides a natural boundary for connecting additional capability, including front-end RTL generators.

\section{Conclusion}
\label{sec:conclusion}

This paper presented \textsc{LATCH}, a two-layer framework separating LLM orchestration from deterministic execution for post-route FPGA frequency improvement across sixteen designs. In a single attempt with seven LLMs, expert recipes passed functional checks in 98.2\% of trials, direct EDA operations in 71.4\% and constructed procedures in 17.9\%, though passing direct operations had a higher median gain of +3.4\% against +3.2\% for expert recipes. Within twelve DCPs per run, mean improvement rose from 10.5\% (sequential) to 16.0\% (branching) and 20.3\% (knowledge base), matching the fitted heuristic's 19.9\%. Per-design best gains spanned 0.4\% to 102.2\%, from iterated recipe selection on twelve designs and one direct-operation attempt on four.

\section*{Acknowledgment}

The authors thank the AMD University Programme for Vivado licenses, and the FPL'26 Agentic FPGA Backend Optimization Competition Organisers for setting the challenge, providing benchmarks, and drafting the baseline software.
We used Claude Sonnet 5, DeepSeek v4 Flash \& v4 Pro, Gemini 3.6 Flash, OpenAI GPT-5.6 Luna, GPT-5.6 Sol, GPT-5.6 Terra as part of the framework development and experimental evaluation; we used Claude Opus 5 to proofread the author-written manuscript text.

\bibliographystyle{IEEEtran}
\bibliography{Sections/References}

\begin{IEEEbiography}
[{\includegraphics[width=1in,height=1.25in,keepaspectratio]{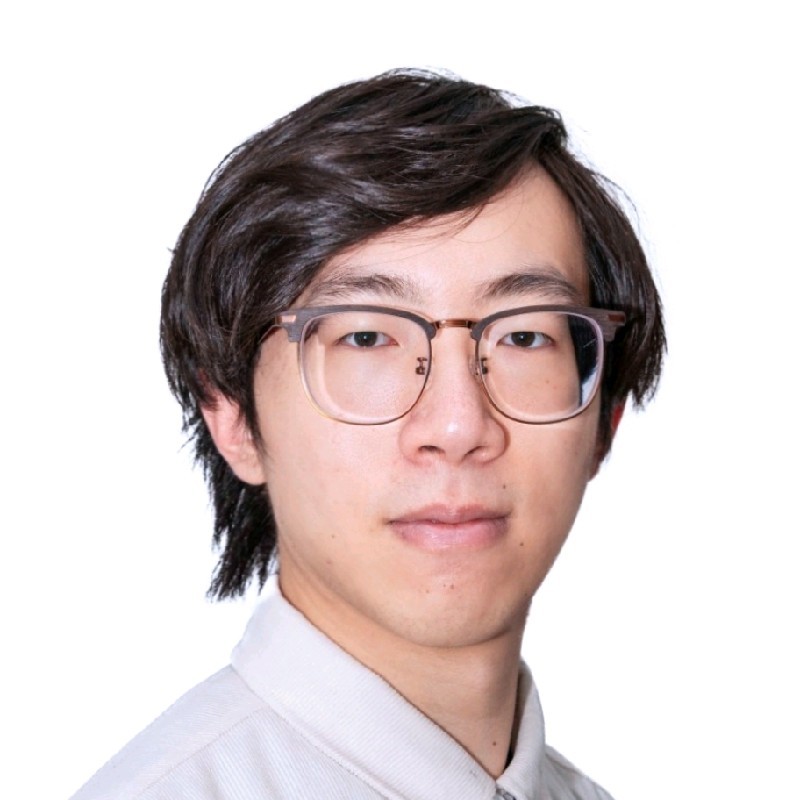}}]{Shu Gu}
received the M.Eng. (Hons.) degree in electronics and computer science from the University of Edinburgh, Edinburgh, U.K., in 2025. He is currently pursuing a Ph.D. degree with the School of Engineering at the same university. His research interests include electronic design automation, FPGA design and timing optimisation, and the application of large language models to hardware design workflows.
\end{IEEEbiography}

\begin{IEEEbiography}
[{\includegraphics[width=1in,height=1.25in,keepaspectratio]{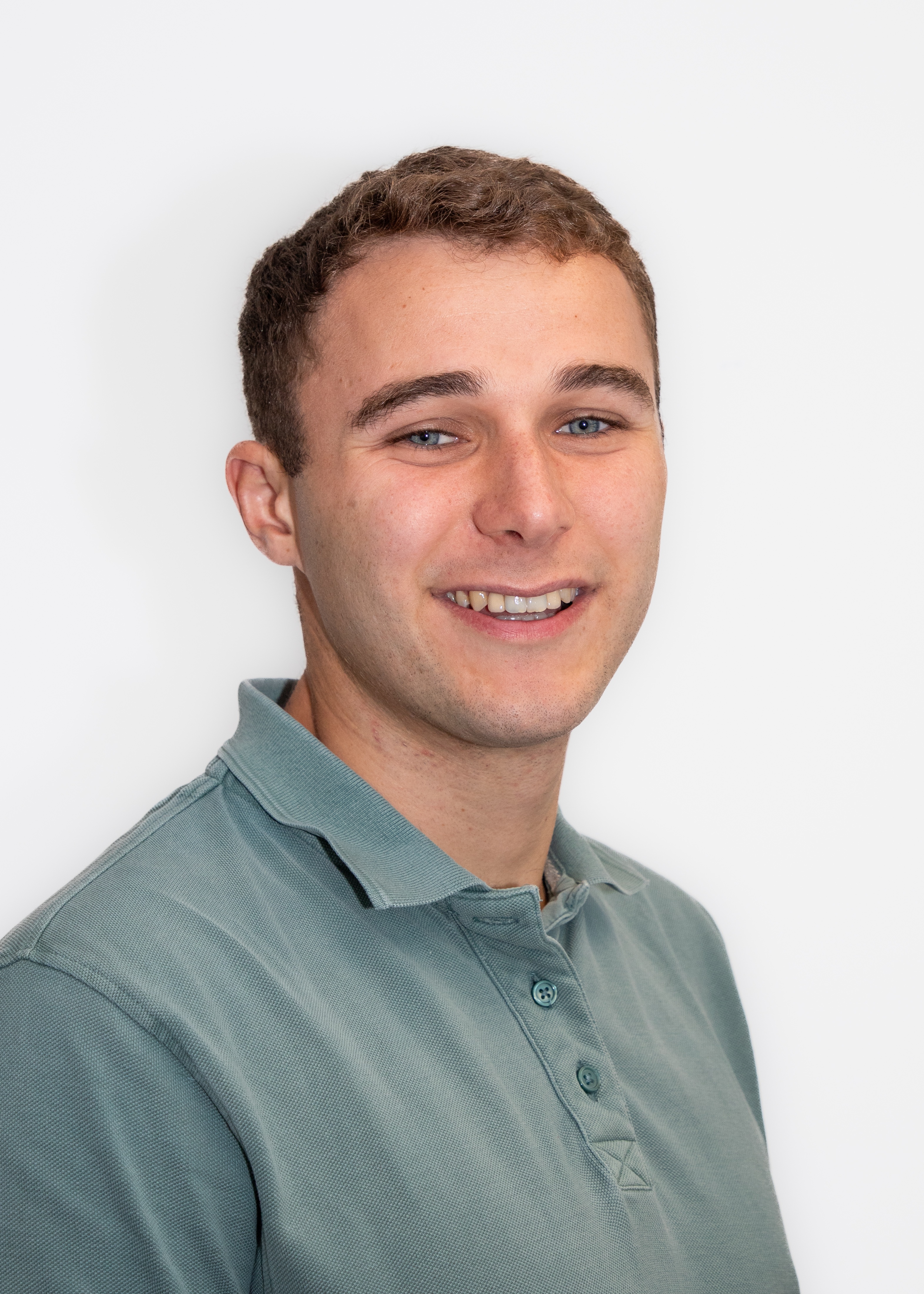}}]{Matthew Youngman} (Graduate Student Member, IEEE) received the M.Eng. degree in general engineering, specialising in electronics, from Durham University, Durham, U.K., in 2024. Following a software engineering role in industry, he is currently pursuing a Ph.D. degree with the School of Engineering, University of Edinburgh, Edinburgh, U.K. His research interests include large language model multi-agent systems for backend electronic design automation, particularly how agents can coordinate deterministic tools across the EDA design flow.
\end{IEEEbiography}

\begin{IEEEbiography}
[{\includegraphics[width=1in,height=1.25in,keepaspectratio]{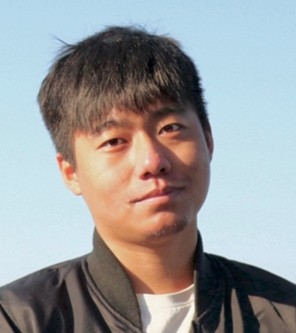}}]{Yuqin Zhao} is a Research Associate at the APRIL AI Hub, University of Edinburgh, U.K. He received his Ph.D. in Electronic and Electrical Engineering from the University of Sheffield, U.K. His research focuses on high-performance heterogeneous computing and AI-assisted hardware design, particularly FPGA/SoC systems, hardware–software task allocation, high-level synthesis, and design automation. His current research interests include generative AI and large language models for electronic design automation, intelligent hardware–software co-design, FPGA acceleration, and AI-assisted hardware verification.
\end{IEEEbiography}

\begin{IEEEbiography}
[{\includegraphics[width=1in,height=1.25in,clip,keepaspectratio]{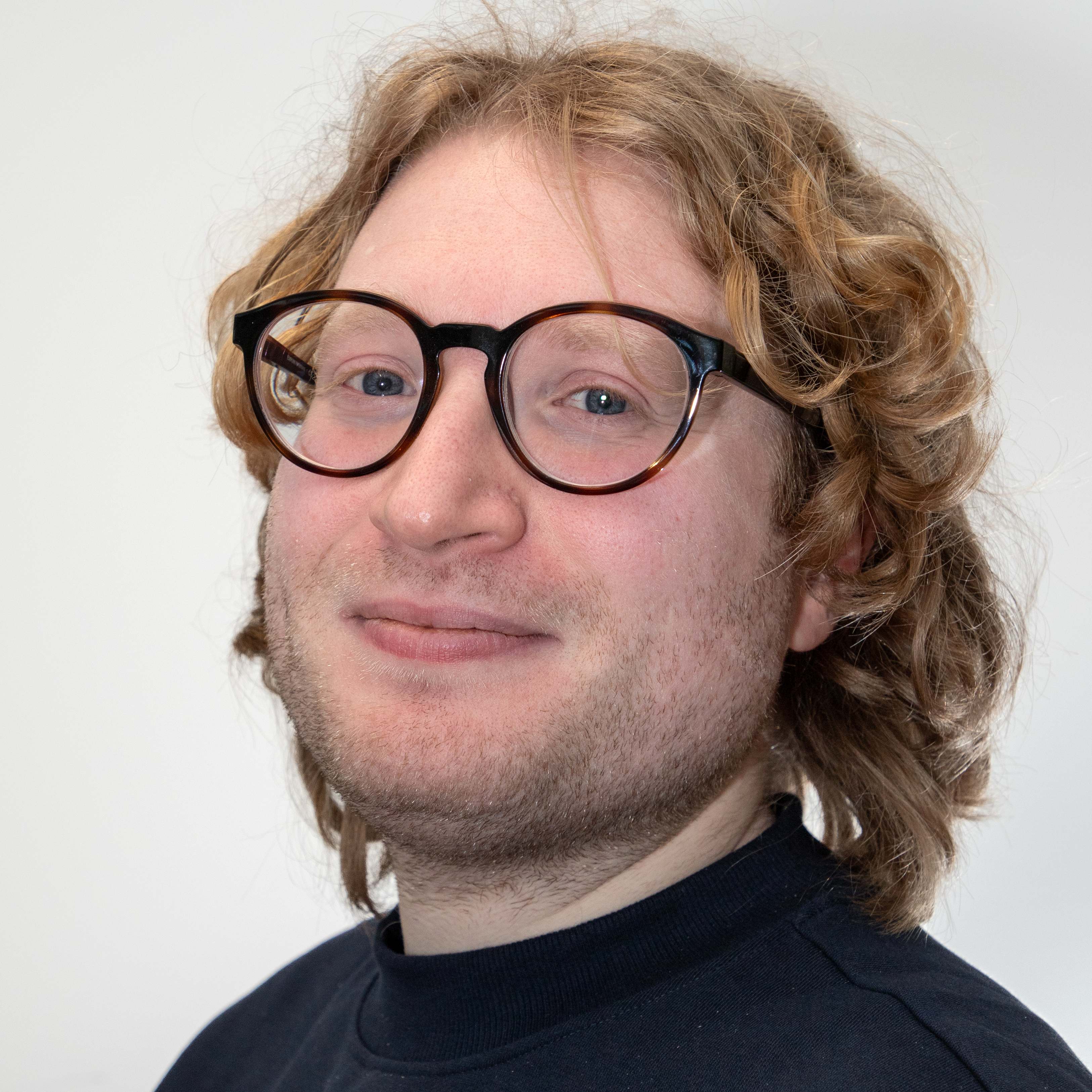}}]{Cristian Sestito}
(Member, IEEE) is a Research Fellow at the Centre for Electronics Frontiers, University of Edinburgh, U.K. He earned his B.Sc. (2016) and M.Sc. (2019) in Electronic Engineering, followed by a Ph.D. in Information and Communication Technologies (2023), from the University of Calabria, Italy. His PhD focused on the design and FPGA implementation of convolutional neural networks, with particular emphasis on dilated and transposed convolutions. During 2021–2022, he was a Visiting Scholar at Heriot-Watt University, Edinburgh, U.K. where he worked on neural network compression techniques. His current research interests include digital AI accelerator design, software simulators for neuromorphic AI, and AI-assisted electronic design automation frameworks.
\end{IEEEbiography}

\begin{IEEEbiography}
[{\includegraphics[width=1in,height=1.25in,keepaspectratio]{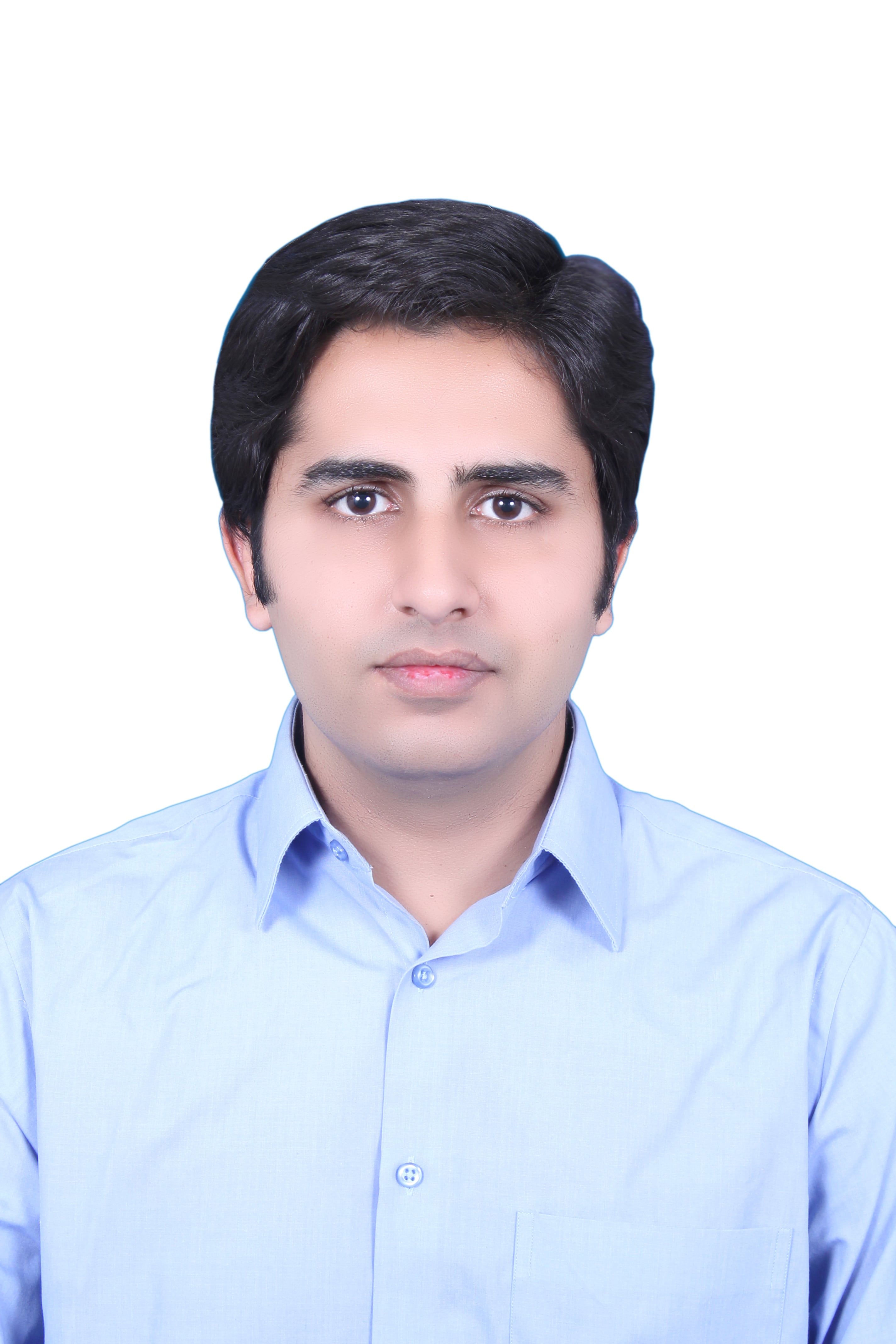}}]{Ali Siddique} is currently a Research Associate in AI for Circuits and Systems Design Automation with the Institute for Integrated Micro and Nano Systems, University of Edinburgh, U.K. Previously, he was a Research Associate with the Advanced Processor Technologies Group, University of Manchester, U.K., where he worked on neuromorphic and edge AI hardware within the EU-funded CONVOLVE project. His research focuses on neuromorphic computing, spiking neural networks, online and on-chip learning, FPGA-based AI accelerators, and hardware-software co-design. His research interests also include agentic AI for hardware design automation and hardware-aware optimization using metaheuristic methods such as Grey Wolf Optimization. He is particularly interested in developing efficient and deployable learning systems that combine algorithmic flexibility with low-cost hardware implementation.
\end{IEEEbiography}

\begin{IEEEbiography}
[{\includegraphics[width=1in,height=1.25in,clip,keepaspectratio]{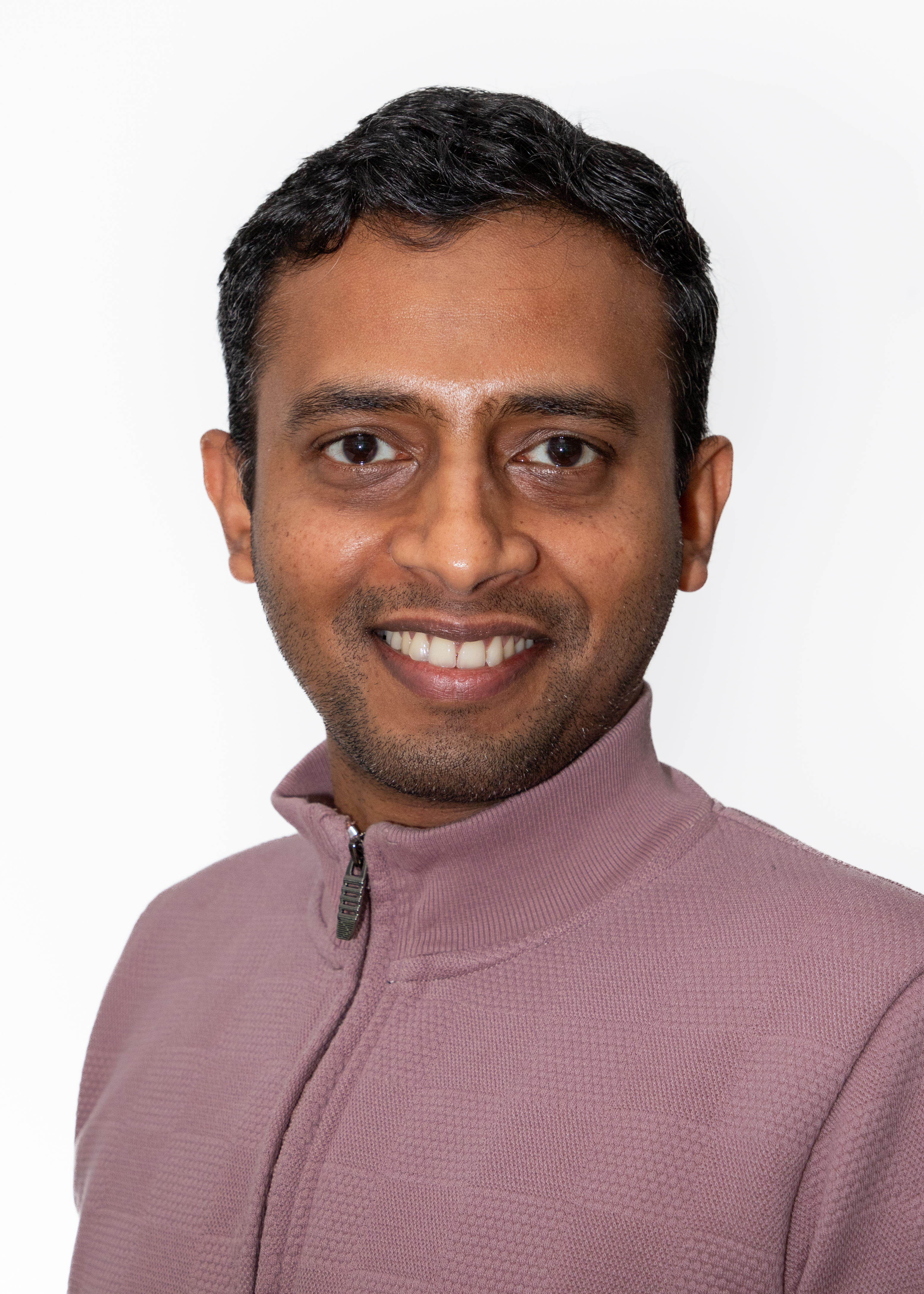}}]{Ramesh Krishnamurthy} is a Research Associate in AI for Electronics Testing and Verification Automation with the APRIL AI Hub at the University of Edinburgh, U.K. He received the B.E. degree in Electronics and Communication Engineering from Visvesvaraya Technological University, India, in 2014, the M.Sc. degree in Embedded Systems from the University of Twente, The Netherlands, in 2021, and the Ph.D. degree in Computer Science from the University of Antwerp, Belgium, in 2025. His doctoral research focused on formal analysis of nonlinear model-predictive control, combining learning-based controller approximation, formal verification, and worst-case execution-time analysis for embedded and safety-critical systems. His current research interests include formal verification, AI-assisted hardware verification, agentic AI for electronic design automation, and trustworthy computing systems.
\end{IEEEbiography}

\begin{IEEEbiography}
[{\includegraphics[width=1in,height=1.25in,clip,keepaspectratio]{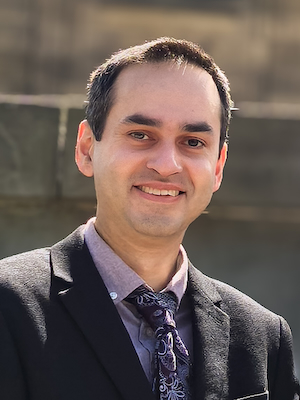}}]{Danial Chitnis}
received an MSc in Microelectronics at Bristol, UK, in 2008, and a DPhil in Engineering Science from the University of Oxford in 2013, and then joined the Department of Biomedical Physics at UCL, London, as a postdoctoral researcher. In 2017, he joined The University of Edinburgh as a Chancellor's Fellow in Electronics. His research focuses on using machine learning and artificial intelligence to improve the design and testing of integrated circuits and electronic systems.
\end{IEEEbiography}

\begin{IEEEbiography}
[{\includegraphics[width=1in,height=1.25in,clip,keepaspectratio]{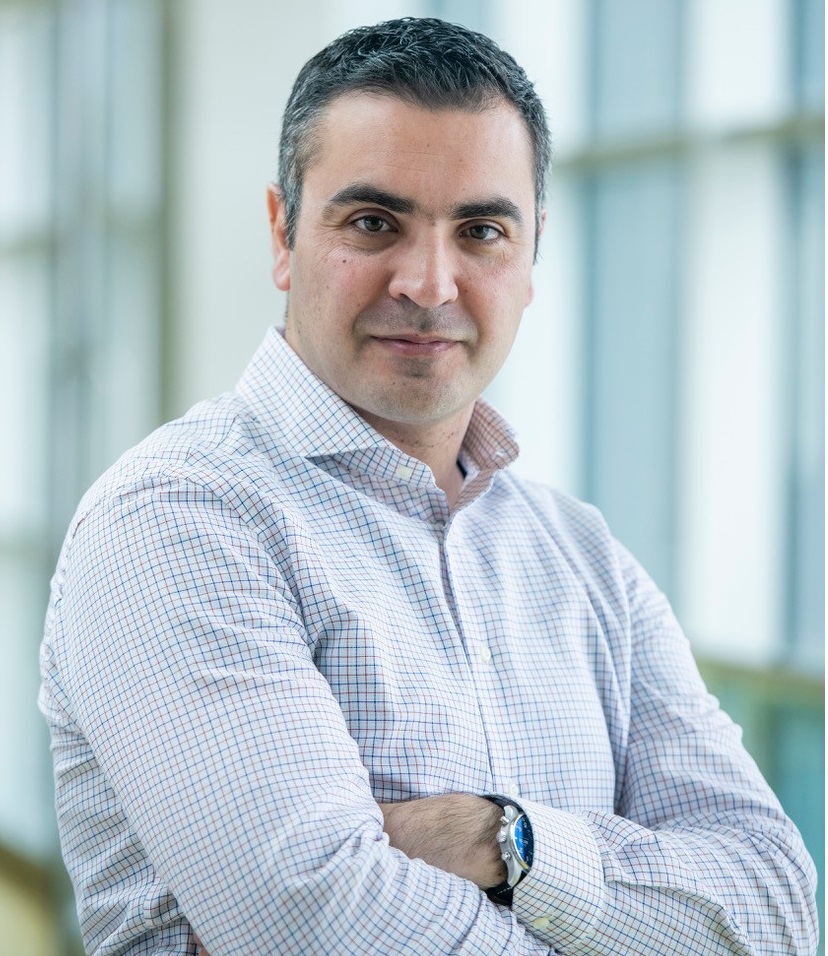}}]{Themis Prodromakis}
(Senior Member, IEEE) received the bachelor’s degree in electrical and electronic engineering from the University of Lincoln, U.K., the M.Sc. degree in microelectronics and telecommunications from the University of Liverpool, U.K., and the Ph.D. degree in electrical and electronic engineering from Imperial College London, U.K. He then held a Corrigan Fellowship in nanoscale technology and science with the Centre for Bio-Inspired Technology, Imperial College London, and a Lindemann Trust Visiting Fellowship with the Department of Electrical Engineering and Computer Sciences, University of California at Berkeley, USA. He was a Professor of nanotechnology at the University of Southampton, U.K. He holds the Regius Chair of Engineering at the University of Edinburgh. He is currently a Royal Academy of Engineering Chair in emerging technologies and a Royal Society Industry Fellowship. His background is in electron devices and nanofabrication techniques. His current research interests include memristive technologies for advanced computing architectures and biomedical applications. He is a fellow of the Royal Society of Chemistry, the British Computer Society, the IET, and the Institute of Physics.
\end{IEEEbiography}

\end{document}